\documentclass[a4paper,11pt]{article}
\usepackage{fullpage}
\usepackage{times}
\usepackage{natbib}

\usepackage{moreverb,url}

\usepackage{hyperref}

\usepackage{graphicx} %
\usepackage{xcolor}
\usepackage{amsmath}
\usepackage{amssymb}  %
\usepackage{underscore}
\usepackage{enumitem}
\usepackage{fancyvrb}
\usepackage{todonotes}
\usepackage{flushend} %

\newcommand{\prism}{\textsc{Prism}}
\newcommand{\bert}{BERT\,2}
\newcommand{\always}{\hbox{\raisebox{-1.2pt}{\large$\Box$}}}

\newcommand{\eventually}{\hbox{\raisebox{0.2pt}{\normalsize$\Diamond$}}}

\newcommand{\until}{~\mathsf{U}}
\newcommand{\G}{\mathsf{G}\,}
\newcommand{\F}{\mathsf{F}\,}
\newcommand{\X}{\mathsf{X}\,}
\newcommand{\U}{\,\until\,}
\renewcommand{\always}{\G}
\renewcommand{\eventually}{\F}

\newcommand{\A}[1]{{\normalfont\textbf{E#1}}}   %

\newcommand{\predcode}[1]{\textrm{#1}}
\newcommand{\pred}[1]{\predcode{#1}}

\definecolor{light-gray}{gray}{0.85}

\newcommand{\pctlstar}{PCTL$^*$}

\newcommand{\new}[1]{{\color{blue}#1\color{black}}}
\newenvironment{newsection}{\color{blue}}{\color{black}}

\newcommand{\newer}[1]{{\color{cyan}#1\color{black}}}
\newenvironment{newersection}{\color{cyan}}{\color{black}}

\renewcommand{\newer}[1]{\new{#1}}

\renewcommand{\new}[1]{#1}
\renewenvironment{newsection}{}{}

\begin{document}

\title{\new{A Corroborative} Approach to Verification and Validation of Human--Robot Teams}


\author{Matt Webster\thanks{Corresponding author. Email: matt@liverpool.ac.uk}~\thanks{Depart\-ment of Computer Science, University of Liverpool, Liverpool, UK}, David Western\thanks{Depart\-ment of Computer Science, University of Bristol, Bristol, UK}, Dejanira Araiza-Illan\footnotemark[3],\\
Clare Dixon\footnotemark[2], Kerstin Eder\footnotemark[3]~\thanks{Bristol Robotics Laboratory, Bristol, UK}, Michael Fisher\footnotemark[2] and \\
Anthony G. Pipe\footnotemark[4]~\thanks{Faculty of Environment and Technology, University of the West of England, Bristol, UK}}
\date{}

\maketitle

\begin{abstract}
We present an approach for the verification and validation (V\&V) of robot assistants in the context of human-robot interactions (HRI), to demonstrate their trustworthiness through \new{corroborative evidence} of their safety and functional correctness.
\new{Key challenges include the complex and unpredictable nature of the real world in which assistant and service robots operate, the limitations on available V\&V techniques when used individually, and the consequent lack of confidence \newer{in} the V\&V results.}
\new{Our approach, called \textit{corroborative V\&V}}, addresses these challenges by combining several different V\&V techniques; in this paper we use formal verification (model checking), simulation-based testing, and user validation in experiments with a real robot.
This combination of approaches allows V\&V of the HRI task at different levels of modelling detail and thoroughness of exploration, thus overcoming the individual limitations of each technique. We demonstrate our approach through a handover task, the most critical part of a complex cooperative manufacturing scenario, for which we propose safety and liveness requirements to verify and validate.
Should the resulting V\&V evidence present discrepancies, an iterative process between the different V\&V techniques takes place until \new{corroboration between  the V\&V techniques} is gained from refining and improving the assets (i.e., system and requirement models) to represent the HRI task in a more truthful manner. \newer{Therefore, corroborative V\&V affords a systematic approach to ```meta-V\&V,'' in which different V\&V techniques can be used to corroborate and check one another, increasing the level of certainty in the results of V\&V.}
\end{abstract}

\section{Introduction}\label{s:introduction}

Robotic assistants that interact with people in an informal, unstructured, and complex manner are increasingly being considered for industrial and domestic domains. 
In manufacturing, the drive towards more flexible production, quality and consistency in the production, and the reduction of tiring and dangerous tasks requires that humans work near robots, or even teach and physically interact with them as co-workers. 

A way to enhance robots, to allow their safe and trustworthy participation in human--robot interactions (HRI), is the incorporation of safety and fault-recovery mechanisms at all levels, \newer{from low-level mechanical systems and basic controllers to higher-level, decision-making systems}~\citep{alamiSafe2006,pipe2011affective}.
For example, restricting motion when near humans has been applied as a low-level safety solution~\citep{Pedrocchi2013}.
However, to allow robot assistants to transition from research laboratories and very limited application scenarios (such as surveillance, transport or entertainment) to the broader domestic and industrial domains, they need to be demonstrably trustworthy~\citep{ROMAN14}. 
Collaborative robots will also need to conform to \new{recent standards, e.g., \citet{iso10218}, \citet{iso13482} and  \citet{iso15066}}.
Thus, HRI requires the development of coherent and credible frameworks for verification and validation (V\&{}V).

A major challenge in V\&V of robot assistants is that no single technique is adequate to cover the whole system in practice. 
``Correct'' functioning in an HRI scenario is likely to depend on precise physical details as well as complex high-level interactions.
Individually, formal methods such as model checking and theorem proving, simulation-based testing, or experiments in real-world scenarios cannot examine the entire state space of the interaction with realistic detail.
The advantages of these techniques --- formal, simulation and experiments --- in terms of \textit{coverability} (i.e., the exploration of the state space such as combinations of human--robot actions, or motion ranges) and \emph{realism} can be exploited when combining them.

Combining V\&V techniques in the HRI domain yields an additional benefit --- trust in the correctness of V\&V results. 
When using a single V\&V technique this is hard to achieve. 
System models used in formal methods or simulation-based testing, and requirements models, are subject to manual input errors, despite efforts in automating translations between models and translations from code to models.
The use of complementary V\&V techniques can highlight discrepancies and help system developers gain confidence in the resulting evidence about safety and liveness requirements. 

\begin{newsection}
\subsection{Our Contribution}

Our contribution, presented in this paper, is twofold: 
\begin{enumerate}
 \item \newer{To propose an approach to the verification and validation of robots
and autonomous systems that allows different V\&V techniques to corroborate one another, and where the outcomes from applying one technique are used to improve the other techniques.
This approach, called \emph{corroborative V\&V}, provides a greater degree of certainty in the V\&V results than would be found in using the V\&V techniques individually.}  
 \item \newer{To demonstrate the effectiveness of corroborative V\&V by applying it to the most critical part of a collaborative manufacturing HRI scenario, the robot-to-human handover task.}
\end{enumerate}
In this paper we combine formal methods, simulation-based testing,
and user validation through experiments with a real robot, in the context of HRI. 
If the evidence agrees when verifying and validating the same requirement through the three techniques, we will be more confident in the results. 
Otherwise, an iterative process is used to refine and improve the truthfulness of the \textit{assets}, the system and requirement models underpinning each technique. 
Hence, corroborative V\&V provides increased confidence in the evidence, compared to using V\&V techniques in isolation.
At the same time, by enabling V\&V to span across multiple levels of detail or abstraction, our approach provides a thorough exploration of the robot's range of behaviours, thus overcoming limitations of individual V\&V techniques. 

\end{newsection}

The proposed approach is exemplified through an object handover task, the most critical component of a cooperative manufacture scenario, for the \bert{} robot~\citep{lenz2010bert2}.
We formulated safety and liveness requirements based on relevant standards.
We then used this case study to show that \new{corroborative V\&V} can provide a higher degree of confidence than when using V\&V techniques in isolation.
The instantiation of our approach for the case study comprises, as V\&V techniques, probabilistic model checking in PRISM~\citep{kwiatkowska11prism}, simulation-based testing in ROS\footnote{\url{http://www.ros.org/}} and Gazebo\footnote{\url{http://gazebosim.org/}}, and an experimental setup in the Bristol Robotics Laboratory (BRL). 

A formal model comprising Probabilistic Timed Automata (PTA) was constructed by hand, representing the HRI. Probabilistic Computation Tree Logic (\pctlstar{})~\citep{kwiatkowska11prism} properties were derived from the requirements, to be verified against the formal model. 
We developed a simulator in ROS--Gazebo, with the real code for the robot and a simulated human co-worker. Tests were derived from model-based and pseudorandom techniques, as in our previous work in~\citep{araizaillan2015,araizaillan2016}, to stimulate the HRI components towards checking the satisfaction of the requirements. Automated checkers implemented as assertion monitors, as described in~\citep{araizaillan2015,araizaillan2016}, were also derived from the requirements and added into the simulator.
Applying the complementary V\&V techniques exposed discrepancies in the resulting evidence, allowing the assets to be examined and refined.
Iterating over this process led to agreement between the three techniques, thus providing greater confidence in the correctness of the resulting evidence and the suitability of subsequent design recommendations.

The paper proceeds as follows. 
Section~\ref{s:approach} presents the \new{corroborative V\&V} approach, outlining the V\&V techniques, their corresponding assets to be developed from the HRI system and its requirements, and their interactions to gain confidence in the resulting evidence. 
We then introduce the case study, the handover task, and the requirements to be verified in Section~\ref{s:handover}. 
Next, we present the instantiation of the proposed \new{corroborative V\&V} approach for the case study in Section~\ref{s:instantiation}, including the development of assets comprising the formal model, the simulator, and the translations of the requirements into logical properties and assertions. 
We present the V\&V results for two of the proposed requirements in Section~\ref{s:verif}, describing in detail the encountered  evidence discrepancies, with the consequent asset refinement and improvement processes until reaching a high degree of \new{corroboration} between the V\&V results. In Section~\ref{s:verification-other-reqs} we then demonstrate V\&V of the remaining requirements using the three V\&V techniques.
Section~\ref{s:discussion} discusses the findings and limitations in the application of the \new{corroborative V\&V} approach to our case study. In Section~\ref{s:relatedwork} we compare our approach to others in the literature, highlighting how \new{corroborative V\&V} provides a novel V\&V framework and complements existing approaches.
Finally, we offer conclusions and directions for future work in Section~\ref{s:conclusions}.

\section{\new{Corroborative V\&V}}\label{s:approach}

\begin{figure}[hbt]
\centering
\fcolorbox{white}{white}{\includegraphics[width=1.0\columnwidth]{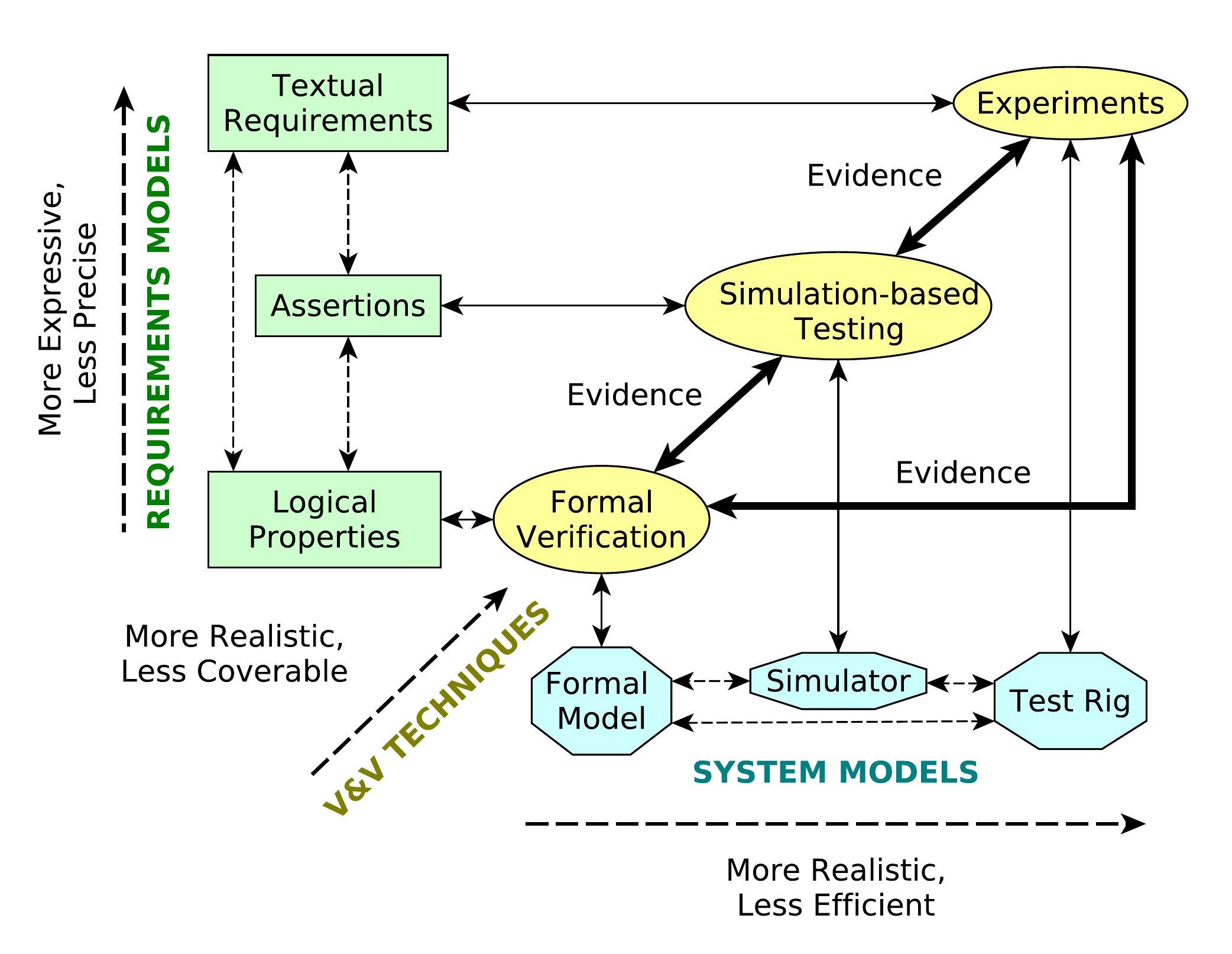}}
\caption{A framework for \new{corroborative V\&V.}}
\label{f:approach}
\end{figure}

As noted in the introduction, \new{corroborative V\&V} provides a thorough exploration of the robot's range of behaviours across multiple levels of abstraction, thus overcoming limitations of individual V\&V techniques.
Our approach to the V\&{}V of human--robot teams is shown in Figure~\ref{f:approach}.
We propose the combined use of multiple techniques to verify and validate robots in HRI tasks, which are shown in ellipses.  
Each technique is underpinned by two \emph{assets}: a requirements model, shown in a rectangle, and a system model, shown in an octagon. In this paper we focus on the use of three particular V\&V techniques, but other methodologies can be integrated into the \new{corroborative V\&V} process if required. This is discussed in more detail in Section~\ref{s:other}.
We introduce the techniques, the assets and the \new{corroborative V\&V} workflows, indicated by the arrows in Figure~\ref{f:approach}, in the following subsections.

\subsection{V\&V Techniques}

\emph{Formal verification} encapsulates a set of mathematical techniques which are used to prove properties about a \emph{formal model} of a system.
Some of the most common formal verification techniques are model checking~\citep{ClarkeMC} and theorem proving~\citep{fitting96first}.
In this paper we use model checking, which lets us verify that formal models (which represent the robot and its non-deterministic environment) satisfy temporal logical properties (derived from requirements) for every possible way in which the models can be executed.
As we examine every possible execution of the formal model, we can demonstrate whether or not the model satisfies the temporal logical properties.

In ``traditional'' model checking, finite state machines are modelled and explored exhaustively in order to determine whether some property holds~\citep{ClarkeMC}. Properties are typically expressed as logical formulae written in a logical language, e.g., linear-time temporal logic (LTL) or computation-tree logic (CTL). The output of a model checker is typically a Boolean value, true or false, indicating whether the model satisfies a given property. In the case where the model does not satisfy the property, an ``error trace'' or counter-example is output describing the sequence of states which led to the violation of the property~\citep{Practical11}. %
Probabilistic model checking, explained further in Section~\ref{s:formalVerification}, extends this method to allow the computation of the probability that a given property will be satisfied.

\emph{Simulation-based testing} involves running a \emph{simulator} under different inputs (or tests), to observe the resulting outputs and determine if the simulated system behaves as intended.
Software and hardware components can be modelled to achieve an appropriate compromise between realism, modelling effort, and computational cost, and real code can be run. 
Nonetheless, the exploration of a system under test is not exhaustive. 
Systematic methodologies to explore the system under test, such as Coverage-Driven Verification (CDV)~\citep{araizaillan2015,araizaillan2016}, should be used to increase efficiency and effectiveness under computational constraints. %
A CDV testing process needs testbench components, including a test generator and a driver, to stimulate the system under test, a coverage collector, to keep track of the V\&V progress, and a checker modelling the requirements and automating the checking~\citep{Piziali2004}. 

\emph{Experiments} are performed within a \emph{test rig} to verify and validate robots interacting in realistic environments with respect to textual requirements. As experiments often involve human volunteers, health and safety assessments and expensive equipment, the number of times that a particular scenario can be examined is often severely limited when compared to simulation or formal verification.
In this paper, experiments are focused towards achieving clear evidence on the principal requirements, as well as to ground the \new{corroborative V\&V} process in reality.

\new{The diagonal axis in Figure~\ref{f:approach} arranges the three techniques based on how realistic and how \emph{coverable} they are, where coverability refers to how much of its asset a technique can analyze. Note that there is generally a trade-off between realism and coverability. Formal verification (e.g., using a model checker) can exhaustively check the entire state-space of a formal model~\citep{ClarkeMC}, while simulation-based testing only samples the state-space of a simulation model. However, a simulation model is able to better account for physical details that are difficult to capture in a formal model, such as physical dynamics, and is therefore able to more realistically model the actual system. Physical experiments are even more realistic, but the number of experiments that can be performed will likely be \newer{significantly lower} than the number of simulations that can be performed, since experiments are more heavily constrained by time and other resources. Additionally, physical experiments can be adversely constrained by ethical or safety concerns which are not an issue in simulation-based testing and formal verification.}

\subsection{V\&V Assets}

In Figure~\ref{f:approach} it is shown that requirements can be modelled in a number of ways. 
\emph{Textual requirements} are the written requirements that describe the desired behaviour of a robot and can also include some assumptions about the human user's behaviour and the environment in which the robot operates (e.g., materials required to complete the task are available at the start). 
Textual requirements are used in experiments to determine whether the robot (i.e., the physical system) satisfies them. Textual requirements for robots are typically based on the needs of the system's users but are increasingly based on legal or ethical frameworks specified by a regulatory or standards authority. For example, \cite{iso15066} defines many safety requirements for collaborative robots.  In practice, verifying a textual requirement in experiments may necessitate refinement of the text with consideration for the actual scenario, to avoid ambiguities.  %

\emph{Assertions} are requirements of a system expressed in an assertion specification language using the syntax of programming languages such as C or Python~\citep{abd}, or as assertion monitors, such as the ones implemented in~\citep{huangRosrv2014,araizaillan2015,araizaillan2016}. 
Assertions are commonly formulated in a precondition-implies-postcondition manner, and can be implemented directly in the code under testing, or within the simulation models. 
Tools are available to convert temporal logical properties into monitors for runtime verification, as in~\citep{Havelund2002,huangRosrv2014}, the latter for testing robots.
The systems under verification are stimulated to attempt triggering the preconditions in the assertions  and consequently a check of their respective postconditions. 
The outcomes of these checks are interpreted to determine if the requirements are satisfied.

A \new{software} \emph{simulator}, usually written in a high-level \new{programming} language, contains models of the robot's behaviour as well as its environment. 
In simulation-based testing, the simulator program is executed multiple times (computation time allowing), to collect information from the assertion checks and the simulation itself.
As mentioned in the introduction, a number of both open-source and proprietary simulation and development frameworks exists in robotics, such as ROS, Player/Stage, Gazebo, V-REP, Webots and others.

\emph{Logical properties} are logical statements, each of which captures one or more requirements of the system using some formal logic. Different logics can be used for different applications; e.g., if we want to capture requirements relating to time we might use Linear Temporal Logic (LTL)~\citep{Practical11}. Alternatively, if we are interested in the probability of the requirement being met then we may use probabilistic computation tree logic (\pctlstar{}).
Formal modelling tools specialize in supporting particular types of formal models and temporal logics, such as \pctlstar{} by PRISM~\citep{kwiatkowska11prism}. 
\emph{Formal models} are discrete computational descriptions of high-level behaviours. Finite State Automata (FSA)~\citep{ClarkeMC} and Probabilistic Timed Automata (PTA)~\citep{prism-manual} are two examples.

Figure~\ref{f:approach} arranges the requirements models in order of how expressive or precise they are.
``Expressivity'' here refers to the breadth of realism that could be referred to in the requirement model, while ``precision'' refers to how specific the expressions may be.
A single requirement may be implemented as assertions in many ways, e.g., according to interpretations by different programmers.
As assertions are based on programming languages whose semantics are more well-defined than natural languages, we consider assertions to be more precise than textual requirements.
Logical properties are, in turn, more precise than assertions and textual requirements, as they have precise, mathematical definitions.
On the other hand, assertions can be more expressive than the logical properties, as they can capture aspects of the system which are difficult to specify at higher levels of abstraction (e.g., physical states that depend on modelled dynamics).
However the assertions are less expressive than the textual requirements: subjective requirements such as user satisfaction are difficult to model in programming languages.
It should also be noted that the more realistic levels of the framework can support a broader set of requirements, since they allow the monitoring of parameters or analysis of components that might not be available in more abstract models. 

Ideally, the assets mentioned before would be generated during the development of the robot itself. 
For example, textual requirements would be developed at the start of the traditional product engineering life cycle and may be based on standards and regulations such as \cite{iso10218} for industrial robots. 
At the next stage in the life cycle the product would be designed. 
Simulation and formal analysis are often used in the hardware and software domains at the design stage in order to gain confidence in the correctness of the design with respect to the specifications. Hence formal models and simulators would be developed. 
This practice can be adopted for the design of robot assistants, as demonstrated by~\citet{Kirwan2013} for autonomous navigation. 
Experiments would be performed during implementation of the robot system in real life HRI, after designing the corresponding setup. 
If it is not possible to develop all assets during the initial development of the human--robot system --- e.g., if the human--robot system has already been developed --- the approach can still be applied. In this case, it is necessary to develop assets based on existing materials.

Re-examining equivalent requirements implemented at different abstraction levels of the framework provides an opportunity to refine individual assets to represent the HRI more accurately and truthfully, making the framework robust with respect to human error and providing a high degree of confidence in the resulting evidence. 
When refining the assets, complexity needs to be carefully managed, e.g., through abstraction. 
Re-modelling formal models and simulators can result in a state-space explosion and a significant increase in time and memory~\citep{Clarke2012}.
As explained previously, each level of our framework represents a different compromise between realism and the coverability of the state space. 
Any decisions affecting the balance of this compromise should be made by those conducting the V\&V.

The bidirectional arrows between the different system models in Figure~\ref{f:approach}, and between the different requirements models, indicate that the development of any of these models may be informed by the equivalent model at another level of abstraction.
Such development may be carried out manually or using some of the techniques mentioned in Section~\ref{s:relatedwork}.
Our framework allows for the incorporation of such techniques to suit the application in question.  

\subsection{\new{Workflows}}\label{s:workflows}

Our approach leaves open the order in which the different V\&V techniques in Figure~\ref{f:approach} should be used.
Such decisions should be made with consideration for the specific HRI application in question.  
Furthermore, these decisions will typically be made in a reactive manner, because insights gained from any of the techniques can lead to modifications in any of the system models or requirements models, necessitating a further stage of V\&V to increase confidence in the results, possibly with a different technique.

For example, we could start with a set of logical properties and a formal model of the robot system.  Formal verification would then be used to verify that the formal model satisfies the logical properties. This process is indicated by the arrows from ``Logical Properties'' and ``Formal Model'' to ``Formal Verification'' in Figure~\ref{f:approach}.
The result of formal verification is evidence that the formal model is correct with respect to the logical properties.
During this V\&V, we may discover that the formal model  does not satisfy a particular property. 
If we trust this V\&V result, modifications to the formal model may be an appropriate way to explore possible design modifications.
The ``Simulator'' and ``Test Rig'' would then need to be updated accordingly, as represented by the bi-directional dashed arrows between system models in Figure~\ref{f:approach}. 
Alternatively, the property violation may be due to an error in the model or in the logical property (i.e., we have incorrectly formalised a requirement).
We may wish to manually revise the properties or formal model (or both) if the fault lies there. This is indicated by the arrows from ``Formal Verification'' back to ``Logical Properties'' and ``Formal Model.''
Similarly, we may wish to gain more confidence in the correctness of the formal model and logical properties by employing one of the other V\&V techniques, before proceeding to modify the real system and the other assets. 

The same requirements, implemented as assertions, could then be monitored during simulation-based testing, providing more V\&V evidence. This technique is indicated by the arrows from ``Simulator'' and ``Assertions'' to ``Simulation-based Testing.'' 
During testing we may find requirement violations, as we did with formal verification, and we have to decide a course of action: revising the relevant assets (e.g., the simulator or the assertions), or proceeding to compare the results with experiments to gain more confidence, if results were similar to formal verification. (The comparison between the outputs of the V\&V techniques is indicated by the bold arrows in~Figure~\ref{f:approach}.)
On the other hand, evidence generated by the simulation may not align with evidence generated by the other V\&V techniques, resulting in a lack of \new{corroboration}.
There are a number of potential causes of such disagreements:
\begin{itemize}
 \item System model inaccuracies. All the V\&V techniques use models of the real-world. The models might have been constructed erroneously, or may be inconsistent with the real world, or relative to one another.
 \item Requirement model inaccuracies. In our approach, the real-world requirements of the system are converted into textual requirements, assertions and properties for V\&V. These requirements models may not have been correctly formulated.
 \item Tool inaccuracies. It is possible that numerical approximations affect the V\&V results. In addition, third party tools can contain bugs that are unknown. 
\end{itemize}
\noindent
We could now proceed to perform ``Experiments.'' 
As before, we may find a problem with the textual requirements or the robot's test rig during experimentation. 
At the same time, the evidence from formal verification and/or simulation-based testing can be compared against the experiment results. 
We may also discover that one of the requirements is satisfied during simulation-based testing or formal verification but not during the experiments. In this case we may need to refine any of the other assets, as explained before.

Careful comparisons must be made between the different representations in order to discover the cause of the conflicts. 
Such comparisons are indicated by the bi-directional bold arrows between ``Formal Verification'' and ``Simulation-based Testing,'' ``Simulation-based Testing'' and ``Experiments,'' and ``Formal Verification'' and ``Experiments,'' respectively, in Figure~\ref{f:approach}.

\section{The BERT Handover Task: A Case Study}\label{s:handover}

\new{Corroborative V\&V can be used to provide a higher degree of confidence in the V\&V evidence than when using V\&V techniques in isolation. In this section, we present an \new{(intentionally) simple} HRI case study to demonstrate this.  The corroborative V\&V of a more complex case study would have been difficult to fully explain within the bounds of this paper. It was thought preferable to cover a simpler scenario in a high level of detail, rather than a more complex HRI scenario in less detail. 
Nevertheless, corroborative V\&V may be applied to more complex scenarios than the one presented here.}

\newcommand{\R}[1]{{\normalfont\textbf{Req. #1}}}

\begin{figure}[hbt]
\centering
\includegraphics[width=0.99\columnwidth]{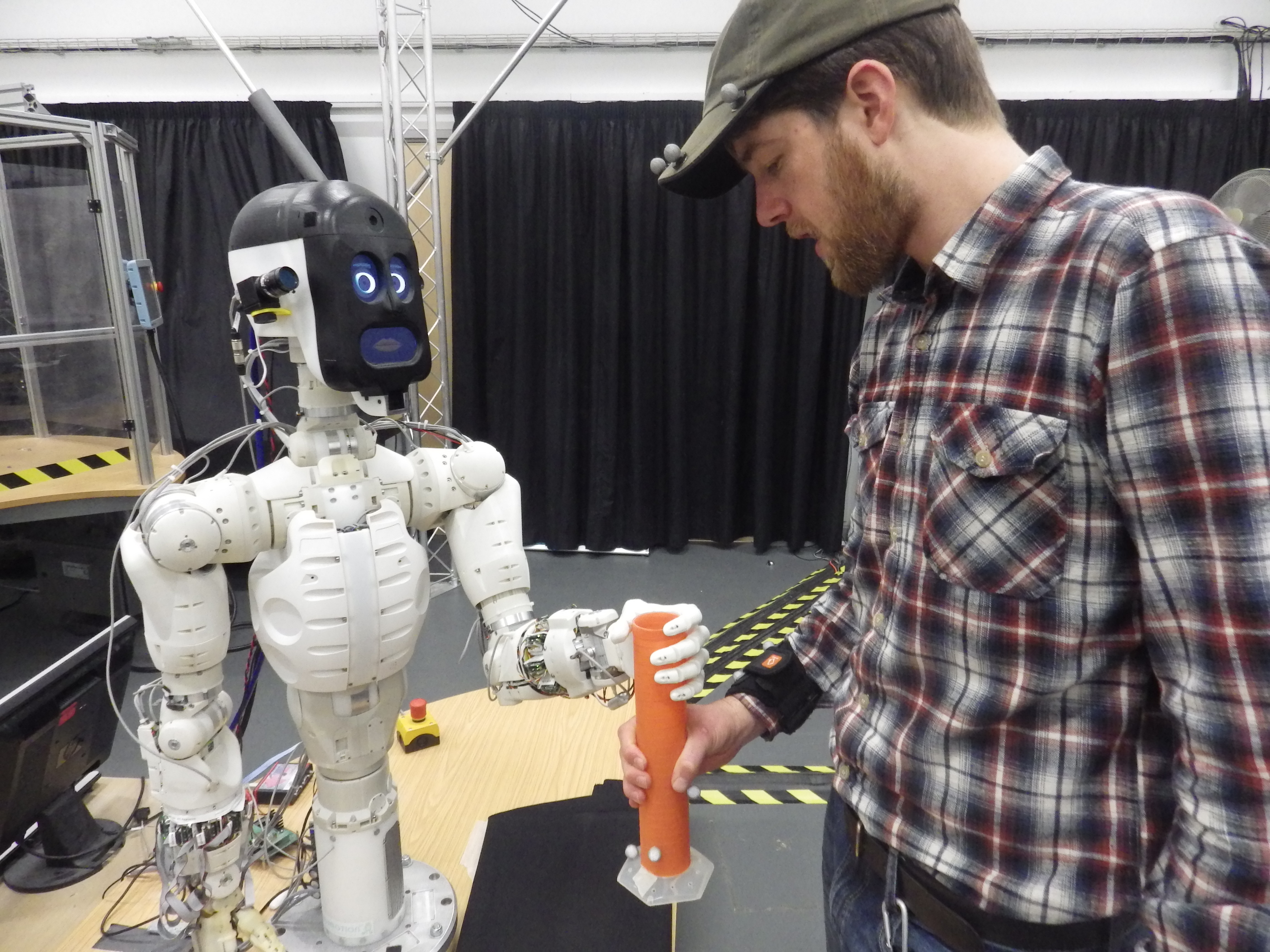}
\caption{\bert{} in the handover task test rig.}
\label{f:bert}
\end{figure}

\newer{Despite its simplicity, our HRI case study concerns robot-to-human handover, the most critical part in a human--robot collaborative manufacturing task.} The case study uses \bert{}, an upper-body humanoid robot designed to facilitate research into complex human--robot interactions, including verbal and non-verbal communication, such as gaze and physical gestures~\citep{lenz2010bert2} (see Figure~\ref{f:bert}).  \bert{}'s software architecture was originally developed using YARP\footnote{\url{http://www.yarp.it}}.  More recently, this system has been wrapped with a ROS interface.

We verify an object \emph{handover} to exemplify our approach, in the context of a broader collaborative manufacture scenario where \bert{} and a person work together to assemble a table~\citep{Lenz2012}.  
In the handover, the first step is an activation signal from the human to the robot. \bert{} then picks up a nearby object and holds it out to the human. The robot announces that it is ready to handover.
The human responds verbally to indicate that they are ready to receive.
(For practical reasons, human-to-robot verbal signals were relayed to the robot by a human operator pressing a key.)
Then, the human is expected to pull gently on the object while looking at it. 
\bert{} then calculates three binary sensor conditions:
\begin{itemize} 
\item Gaze: The human's head position and orientation relative to the object are tracked using the Vicon\textsuperscript{\textregistered} motion-tracking system for an approximate measure of whether he/she is looking at the object.
\item Pressure: Changes in the robot's finger positions are sensed to detect whether the human is applying pressure to take the weight of the object.
\item Location: The Vicon\textsuperscript{\textregistered} motion-tracking system is used to determine whether the human's hand is located on the object.
\end{itemize}
The sensor conditions must be calculated within a time threshold for \bert{} to determine if the human ``is ready.'' The robot should release its grip on the object if all three conditions are satisfied. Otherwise, the robot should terminate the handover and not release the object.
The human may disengage and the robot can timeout, which would cancel the remainder of the handover task.
The sensors are not completely accurate and may sometimes give incorrect readings.

\new{\subsection{System Requirements}}

A safety requirement ensures that ``nothing bad happens,'' whereas a liveness requirement ensures that ``something good happens eventually'' or inside a threshold of time for practical reasons (e.g., in simulation). 
The requirements for any HRI task depend on its safety and functional context. 
For example, in our case study the robot would need to achieve a particular handover success rate threshold to keep up with manufacturing throughput or avoid unacceptable damage costs, as per the users' requirements.
We considered two different thresholds for our first functional requirement, based on estimates of acceptable productivity in two different settings. 
The first threshold is considered for deployed use in a hypothetical manufacturing environment. 

\begin{itemize}[leftmargin=2cm]
\item [\R{1a}:] At least 95\% of handover attempts should be completed successfully.
\end{itemize}
In a research and development environment, a lower threshold may be considered satisfactory to provide proof-of-concept, showing that the system works most of the time.

\begin{itemize}[leftmargin=2cm]
\item [\R{1b}:]\label{req1} At least 60\% of handover attempts should be completed successfully.
\end{itemize}
The following requirements were chosen to illustrate our approach, inspired by~\citet{Grigore2011} and drawing from standards \cite{iso10218} for industrial robots, \cite{iso13482} for personal care robots, and \cite{iso15066} for collaborative robots: %

\begin{itemize}[leftmargin=2cm]
\item [\R{2}:]\label{req2} If the human is not ready, the robot shall not hand over the object.

\item [\R{3}:]\label{req3} If the human is ready, the robot shall hand over the object.

\item [\R{4}:]\label{req4} The robot always reaches a decision within a threshold of time.

\item [\R{5}:]\label{req5} The robot shall always either time out, decide to release the object, or decide not to release the object.

\item [\R{6}:]\label{req6} The robot shall not close its hand when the human is too close.

\item [\R{7}:]\label{req7} The robot shall start in restricted speed.

\item [\R{8}:]\label{req8} If the robot is within 10 cm of the human, the robot's hand speed is less than 250 mm/s.
\end{itemize}
These requirements are ambiguous in terms of how they are assessed over the available system information, reflecting the generality of the standards and the shortfalls of using natural language when first establishing requirements. 
In order to verify and validate them, we need to interpret them in terms of available variables and system behaviours according to the assets. 

\section{\new{Corroborative V\&V} of the Case Study}\label{s:instantiation}

\label{s:plan}

After establishing the system's requirements, we developed a plan for the application of \new{corroborative V\&V} to the case study. We chose to focus on a ``typical use case'' for the handover task, in which the human has a working familiarity with the robot and intends to complete the task successfully.
Any of the requirements may be used as bases for comparison between techniques used, provided that the requirement may be modelled at all levels of abstraction.
We chose to focus on our principal functional requirements (\textbf{Reqs.\ 1a--b}, concerning the handover success rate) as a first basis for failure finding and to refine the assets if necessary. 
The handover success rate could be expected to be sensitive to a wide range of foreseen and unforeseen events. 
As a scalar measure, it allows evidence from the V\&V techniques to be compared in a quantitative manner,  whereas comparisons of Boolean results may be insensitive to important modelling discrepancies.
 
After focusing on \textbf{Reqs.\ 1a--b}, we proceed to verify the remaining requirements (\textbf{Reqs.\ 2--8}), identifying any further need to improve assets or the system itself.
The V\&V of the full set of requirements provides a more comprehensive evaluation of the system's requirement satisfaction, whilst facilitating the evaluation of the benefits of combining individual V\&V techniques to complement one another.

As mentioned previously in Section~\ref{s:workflows}, \new{corroborative V\&V} will be carried out in a reactive manner according to the resulting evidence.
In terms of the order in which we applied the V\&V techniques, we chose to begin with a comparison of formal verification and simulation-based testing for \textbf{Reqs.\ 1a--b}, to acquire as much insight as possible into the system and our modelling assumptions before committing resources to more expensive physical experiments.
The subsequent stages of V\&V and asset modification, explained later in Section~\ref{s:verif}, were conducted with the aim of achieving agreement on the handover success rate (\textbf{Reqs.\ 1a--b}) that was \new{corroborated} by all three V\&V techniques.

In order to apply our approach to the \bert{} handover scenario, it was necessary to implement each element in Figure~\ref{f:approach}.
Appropriate tools for formal verification and simulation-based testing were selected first.
Requirements models were then translated from the textual requirements in Section~\ref{s:handover}, and system models were constructed to reflect the physical system.
We developed relevant assets for a chosen set of tools comprising the probabilistic model checker PRISM, ROS--Gazebo and a CDV testbench for simulation-based testing, and experiment designs at the BRL. 
We detail the development of these components in the following subsections.

\subsection{Formal Verification}\label{s:formalVerification}

We chose \prism{}, a probabilistic symbolic model checker~\citep{kwiatkowska11prism}, for the formal verification component. 
In \prism{}, probabilistic systems can be modelled as discrete- and continuous-time Markov chains, Markov decision processes and Probabilistic Timed Automata (PTA). 
In \prism{} models, transitions between states can be annotated with probabilities. 
The models consist of a set of modules, each representing a different process within the system being modelled. 
Modules are executed concurrently. Each module consists of a number of variables along with transition rules for updating those variables according to preconditions. Communication between modules is made possible by reading globally-accessible variables and by synchronisations between transitions in different modules. Execution of a \prism{} model starts from an initial state (of which there can be many) and advances by application of transitions whose preconditions have been satisfied. These transitions then update the state of the model. This continues until a fixed point is reached when it is no longer possible to update the state~\citep{prism-manual}.

Properties to verify can be expressed in a probabilistic logic such as probabilistic computation tree logic (\pctlstar{})~\citep{baier08principles}. Rather than outputting a Boolean value, \prism{} can be used to output a probability that a given property holds for some sequence of states, or \emph{path}, through a model~\citep{prism-manual}.
\prism{} has been used to model and verify a range of probabilistic systems such as security protocols~\citep{Duflot2013}, biological systems~\citep{Konur2015}, robots and multi-robot systems~\citep{Konur2012,Llarena2012}. 

\subsubsection{Formal Model}\label{s:formalmodel}

\begin{figure}
\centering
\caption{\new{Inter-module communication within the \prism{} formal model.}}
\fcolorbox{white}{white}{\includegraphics[width=.6\columnwidth]{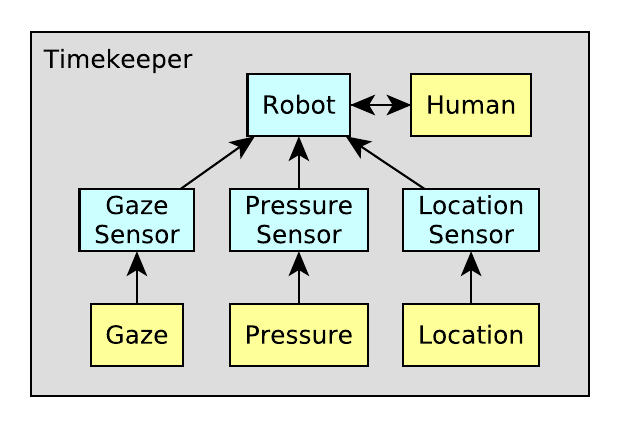}}
\label{f:prism-model}
\end{figure}

The \prism{} model of the handover task consists of nine different modules: the \textit{human}, the human's \textit{gaze}, hand \textit{pressure} and \textit{location}, the \textit{robot} (representing \bert{}), \bert{}'s \textit{gaze}, \textit{pressure} and \textit{location sensors}, as well as a \textit{timekeeper} module which keeps track of time elapsed in the model. \new{Figure~\ref{f:prism-model} shows how the different modules within the \prism{} formal model communicate. There are four modules which model the human's behaviour: Human, which models the human's decision-making and communications with the robot; and gaze, pressure and location, which model the human's gaze, hand pressure and hand location. Four modules model the \bert{} robotic system: Robot, which models the robot's decision-making and communication with the human, and gaze sensor, pressure sensor and location sensor, which model sensors which are tracking the human's gaze, hand pressure and hand location. The Timekeeper module monitors all of the other modules to measure time elapsed.}

The model consists of around 300 lines of \prism{} code and is therefore too long to reproduce in this paper. \new{We could represent the PRISM modules as diagrammatic state transition systems; however, due the large number of states of each module such
diagrams are hard to read. Therefore, for  illustrative purposes the Robot, Human and Timekeeper modules are shown in Figures~\ref{f:prism-code-human}, \ref{f:prism-code-robot} and \ref{f:prism-code-timekeeper} respectively.} \new{Additionally, the full code for the PRISM model is available online~\citep{webster16prismcode}.}

The \prism{} code can be interpreted as follows. The first line in the Human module defines the start of the module. The second line defines a module variable, ``humanState,'' which is an integer in the range 0 to 99. Its initial value is set to ``start'' which is the name of a constant integer set outside the module:

\begin{scriptsize}
\begin{Verbatim}[tabsize=4]
const int start = 0;
\end{Verbatim}
\end{scriptsize}
\begin{figure}
\centering
\caption{The Human module written in \prism{}.}
\begin{scriptsize}
\begin{Verbatim}[tabsize=4,fontfamily=helvetica,numbers=right,numbersep=-6pt,commandchars=\\\{\}]
module human
	humanState : [0..99] init start;
	[activateRobot] humanState=start \texttt{->} (humanState'=activatedRobot);
	[tick] humanState=activatedRobot \texttt{->} (humanState'=activatedRobot);
	[informHumanOfHandoverStart] humanState=activatedRobot \texttt{->}
	  (humanState'=responding);
	[humanIsReady] humanState=responding \texttt{->} (humanState'=setGPL);
	[tick] humanState=setGPL \texttt{->} pDisengages: (humanState'=offTask) +
	  pStaysOnTask: (humanState'=setGPL);
endmodule
\end{Verbatim}
\end{scriptsize}
\label{f:prism-code-human}
\end{figure}
\begin{figure}
\centering
\caption{The Robot module written in \prism{}.}
\begin{scriptsize}
\begin{Verbatim}[tabsize=4,fontfamily=helvetica,numbers=right,numbersep=-6pt,commandchars=\\\{\}]
module robot
	robotState: [1100..1199] init waiting;
	handContents : [0..1000] init nothing;
	testedTimeout: bool init false;
	GPLWasOk: bool init false;
	[activateRobot] robotState=waiting \texttt{->} (robotState'=
	  moveHandToObjectLocation);
	[movingHand] robotState=moveHandToObjectLocation \texttt{->} (robotState'=
	  graspObject) & (handContents'=leg);
	[graspingObject] robotState=graspObject \texttt{->} (robotState'=
	  moveHandToHandoverLocation);
	[informHumanOfHandoverStart] robotState=moveHandToHandover-
	  Location \texttt{->} (robotState'=informedHumanOfHandoverStart);
	[humanIsReady] robotState=informedHumanOfHandoverStart \texttt{->}
	  (robotState'=waitForGPLUpdate);
	[GPLOkSet] robotState=waitForGPLUpdate & humanState=setGPL &
	  gazeSensorState=gazeOk & pressureSensorState=pressureOk &
	  locationSensorState=locationOk \texttt{->}
	  (robotState'=GPLOk) & (GPLWasOk'=true);
	[tick] robotState=waitForGPLUpdate & humanState=setGPL &
	  (gazeSensorState=gazeNotOk \texttt{|} pressureSensorState=pressureNotOk \texttt{|}
	  locationSensorState=locationNotOk) \texttt{->} (robotState'=waitForGPLUpdate);
	[tick] robotState=waitForGPLUpdate & (humanState=tired \texttt{|}
	  humanState=bored \texttt{|} humanState=offTask) \texttt{->} (robotState'=
	    interactionDone);
	[tick] robotState=GPLOk \texttt{->} (handContents'=nothing) &
	  (robotState'=handoverSuccessful);
endmodule
\end{Verbatim}
\end{scriptsize}
\label{f:prism-code-robot}
\end{figure}
\begin{figure}
\centering
\caption{\new{An excerpt from the Timekeeper module written in \prism{}.}}
\begin{newsection}
\begin{scriptsize}
\begin{Verbatim}[tabsize=4,fontfamily=helvetica,numbers=right,numbersep=-6pt,commandchars=\\\{\}]
module timekeeper
	time: [0..timeMax] init 0;  // time in tenths of a second
	objectReleaseTimer: [0..ortMax] init 0;
	[activateRobot] !GPLWasOk & time\texttt{<=}(timeMax-40) \texttt{->} (time'=time+40);
	[informHumanOfHandoverStart] !GPLWasOk & time\texttt{<=}(timeMax-60) \texttt{->} 
	  (time'=time+60);
	[movingHand] !GPLWasOk & time\texttt{<=}(timeMax-100) \texttt{->} (time'=time+100);
	[graspingObject] !GPLWasOk & time\texttt{<=}(timeMax-50) \texttt{->} (time'=time+50);
	[humanIsReady] !GPLWasOk & time\texttt{<=}(timeMax-10) \texttt{->} (time'=time+10);
	[senseGaze] !GPLWasOk & time\texttt{<=}(timeMax-1) \texttt{->} (time'=time+1);
	[sensePressure] !GPLWasOk & time\texttt{<=}(timeMax-1) \texttt{->} (time'=time+1);
	[senseLocation] !GPLWasOk & time\texttt{<=}(timeMax-1) \texttt{->} (time'=time+1);
	[tick] !GPLWasOk & time\texttt{<=}(timeMax-1) \texttt{->} (time'=time+1);
	[GPLOkSet] !GPLWasOk & time\texttt{<=}(timeMax-1) \texttt{->} (time'=time+1) & 
		(objectReleaseTimer'=0);
	[activateRobot] GPLWasOk  & time\texttt{<=}(timeMax-40) & 
		objectReleaseTimer\texttt{<=}(ortMax-40) \texttt{->} (time'=time+40) & 
		(objectReleaseTimer'=objectReleaseTimer+40);
	[informHumanOfHandoverStart] GPLWasOk  & time\texttt{<=}(timeMax-60) & 
		objectReleaseTimer\texttt{<=}(ortMax-60) \texttt{->} (time'=time+60) & 
		(objectReleaseTimer'=objectReleaseTimer+60);
	// (The rest of the file has been removed for the sake of brevity.)
endmodule
\end{Verbatim}
\end{scriptsize}
\end{newsection}
\label{f:prism-code-timekeeper}
\end{figure}
\newcommand{\texttts}[1]{{\scriptsize\texttt{#1}}}
Lines 3--9 are transition rules, which determine how the state of the Human module changes over time. For example, the first rule (see line 3) says that if the human is in the state called ``start,'' then the state is updated to ``activatedRobot.'' In other words, the first thing the human does in this scenario is to activate the robot for the handover task. The rule also contains a \emph{synchronisation} label, ``activateRobot,'' which means that this transition must occur at the same time as all other transitions with the same label. \new{In this case, the only other module containing this label is the Timekeeper module \new{(Figure~\ref{f:prism-code-timekeeper})}, as the synchronisations in this model are used primarily to keep track of how much time has elapsed.} Another feature of \prism{} is probabilistic non-determinism, which can be seen in lines 8--9 of the Human module, in which the human may disengage from the handover task with a probability set by \texttts{pDisengages} or remain engaged with a probability set by  \texttts{pStaysOnTask}.  These are modelled as two constant double-precision floating point numbers:

\begin{scriptsize}
\begin{Verbatim}[tabsize=4]
const double pDisengages  = 0;
const double pStaysOnTask = 1-pDisengages;
\end{Verbatim}
\end{scriptsize}
\new{For our case study, the probability that the human disengages (i.e., becomes bored or distracted) is set to zero as we are examining the typical use-case in which the human is always focused on the task. Similarly, we assume that the human's gaze, hand pressure and location are always within acceptable bounds for the handover task, i.e., the probabilities that these are acceptable is set to 1.0. In this model we are primarily concerned with the robot's reliability, so we assume that the human is completely reliable and engaged with the task at hand. Note that these probabilities could be set differently if, for instance, we wanted to incorporate the human's tiredness level into the model, \newer{or if we wanted to specify that the person's interest in the task may waver and affect their gaze and hand pressure/location.}

Clearly, the Human module only captures the tiny fragment of human behaviour which is relevant to the handover sub-task. In more complex HRI scenarios, the human module may have to be much more complex. Indeed, it is extremely unlikely that a PRISM module will ever be able to capture the full complexity and  nuance of human behaviour. However, it is still desirable, and in fact necessary, for V\&V to model the human's interactions with the robot, even if the model is abstract and coarse-grained.}

Real-world sensors do not work perfectly, and this is reflected in the formal model. As a result, it is possible that the handover task will not always complete successfully.
 The gaze sensor  reports that the human is looking at the object only 95\% of the time. The rest of the time the sensor reports (incorrectly) that the human is not looking at the object. When the gaze sensor reports correctly that the human's gaze is okay, the gaze sensor has reported a ``true positive.'' When the gaze sensor incorrectly reports that the human's gaze is not okay, we call this a ``false negative.'' Similarly, the gaze sensor may correctly report that the person is not looking at the object (a true negative, with probability 95\% also) or may incorrectly report that the person is looking at the object (a false positive). 
 
 \new{The part of the formal model which handles the gaze, pressure and location sensor states can be seen in lines 16--22 of Figure~\ref{f:prism-code-robot}.} Note that true positives and their corresponding false negatives are mutually exclusive, and therefore $P(\pred{false\_negative}) = 1-P(\pred{true\_positive})$. The same is also true of true negatives and false positives:

\begin{scriptsize}
\begin{verbatim}
const double pGazeTP			= 0.95;
const double pGazeFN			= 1-pGazeTP;
const double pGazeTN			= 0.95;
const double pGazeFP			= 1-pGazeTN;
\end{verbatim}
\end{scriptsize}
The pressure and location sensors are given the same probabilities of 95\% for true positives/negatives and 5\% for false negatives/positives. With no experimental results or hardware specifications to refer to, it was assumed that sensors would be accurate ``most of the time.'' A reliability of 95\% was therefore chosen as a first estimate.

\subsubsection{Logical Properties}

Logical properties, representing requirements, were expressed in terms of \pctlstar{}.  We use the following \pctlstar{} symbols~\citep{prism-manual}: $\neg p$ meaning that $p$ is not true, $p\wedge q$ meaning that both $p$ and $q$ are true, $p\vee q$ meaning $p$ and/or $q$ is true, $p \implies q$ meaning that if $p$ is true then  $q$ is true, $\eventually p$ meaning that eventually $p$ will be true, $\always p$ meaning that $p$ is always true from now on, $\X p$ meaning that $p$ is true in the next state and $p\until~q$ meaning that $p$ is true until $q$ is true. $P(q)$ denotes the probability of $q$ being true in the initial state.

For example, consider \newer{\R{3}}: ``Once the human is ready, \bert{} will hand over the object.'' This requirement can be implemented as a temporal logical formula:

\begin{scriptsize}
\begin{equation}\label{p:time1}
\begin{array}{l}
\G (\pred{robotState=GPLOk} \implies~\F\pred{robotState=handoverSuccessful})
\end{array}
\end{equation}
\end{scriptsize}%
which reads ``it is always the case that if gaze, position and location are correct, then eventually the handover is successful (i.e., the object is released to the human).'' We can then find the probability of this formula being true on any given path through the state space. We do this by forming a property in probabilistic computation tree logic (specifically, PCTL*) which can be analysed using a probabilistic model checker like PRISM:

\begin{scriptsize}
\begin{equation}\label{p:time}
P^{=?} \left(
\begin{array}{l}
\G (\pred{robotState=GPLOk} \implies~\F\pred{robotState=handoverSuccessful})
\end{array}\right) 
\end{equation}
\end{scriptsize}%
Using the operation $P^{=?}(f)$ tells the model checker that we want to find out the probability of the formula $f$.

Another requirement, \newer{\R{1a}}, is that the probability of completion of the handover task should be greater than 95\%. This can be rephrased as, ``the success rate of the handover task is at least 95\%.'' This can be formulated as a property in PRISM as follows:

\begin{scriptsize}
\begin{equation}
P \left(
\begin{array}{l}
\F \pred{robotState=handoverSuccessful} \\
\end{array}\right) \ge 0.95
\end{equation}
\end{scriptsize}%
This property states that the probability that the robot will eventually release the object is at least 0.95, or 95\%.

Note that the translation of textual requirements into logical properties is not direct, since there might be different interpretations depending on the available variables, probabilities, and so on. Hence, this translation process carries the potential for misinterpretation. For example, in the properties above, ``$\pred{handoverSuccessful}$'' is used as a synonym for ``object handed over,'' which may not be correct in all cases (e.g., the human may drop the object after release).

The full code for the \prism{} models and properties used in this paper can be found online~\citep{webster16prismcode}.

\subsection{Simulation-Based Testing}\label{s:simulationBasedTesting}

\begin{figure}[hbt]
\centering
\includegraphics[width=0.99\columnwidth]{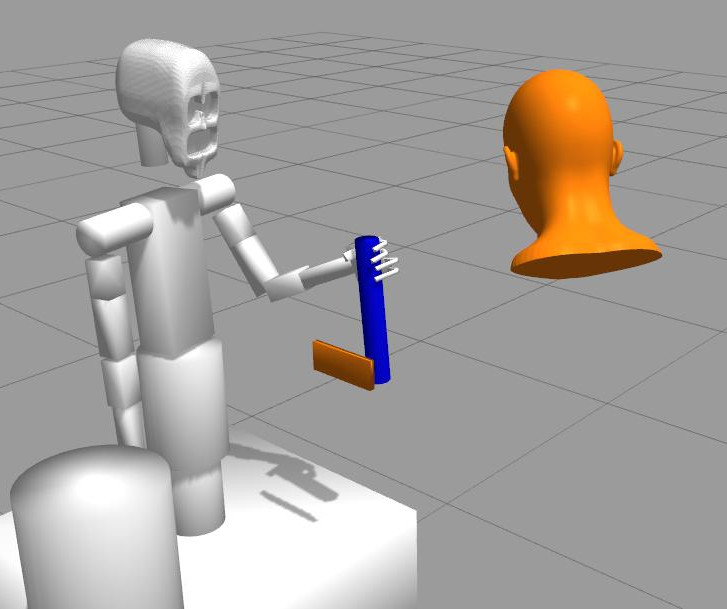}
\caption{Screenshot of the simulated handover task. The human head and hand are represented in orange. The object to be handed over is shown in blue.}
\label{f:sim}
\end{figure}

A simulator for the handover task was implemented in the ROS framework for robot code development and the Gazebo simulator.
Among Gazebo's features are support for 3D graphics rendering and various physics engines (including ODE\footnote{\url{http://www.ode.org}}, used in this paper).
Although now available as a standalone Ubuntu Linux package, Gazebo was originally developed as a ROS package and retains its compatibility with ROS.
A URDF (Universal Robot Description Format) file, used in ROS to describe the kinematic structure of the robot, actuators, and sensors, can simply be extended to describe parameters used by the physics engine such as inertial properties and friction coefficients.
This compatibility allows the same control code to be used in simulation and in the actual robot, providing consistency between simulations, experiments, and deployed use. A screenshot of the ROS/Gazebo simulation can be seen in Figure~\ref{f:sim}.

For the simulator, additional ROS nodes were constructed in Python to simulate \bert{}'s sensor systems and embedded actuation controllers.
The pre-existing URDF file describing \bert{} was extended as described previously for use in Gazebo.
The simulated human behaviour was controlled by a ROS node written in Python driving a simplified physical model of the head and hand.

A testbench was incorporated into the simulator. 
The testbench comprised a test generator, a driver, a checker and a coverage collector. 
Achieving the exploration of meaningful and interesting sequences of behaviours from the robot and its environment in an HRI task is a challenging task. 
For this reason, we stimulate the robot's code in the simulation indirectly through stimulating its environment (e.g., the person's behaviour) instead, and we use a combination of model-based and pseudorandom test generation.
Also, to alleviate the complexity of generating and timing different types of system inputs, the test generator is based on a two-tiered approach~\citep{araizaillan2016} where an abstract test is generated first and then concretized by instantiating low-level parameters.
The high-level actions of the human in the simulator include sending signals to the robot or setting abstract parameters for gaze, location and pressure. 
Low-level parameters include the robot's initial pose and the poses and force vectors applied by the human during the interaction.
For example, we computed an abstract test of high-level actions for the human, by exploring the model in UPPAAL\footnote{\url{http://www.uppaal.org}}, so that the robot was activated (sending a signal to activate the robot and wait for the robot to present the object), the gaze, pressure and location sensor readings were correct (set gaze, pressure and location to mean ``ready''), and the robot released the object. 
This allowed testing \R{3}, ``If the human is ready, \bert{} should hand over the object.'' 

The driver distributed the test components in the simulator.
A self-checker --- i.e., automated assertion monitors --- was added according to the requirements, described in more detail in the following subsection.
Finally, a coverage collector gathered statistics on the triggering of the assertion monitors. 
The simulator code is available online\footnote{\url{https://github.com/robosafe/testbench_ABV}}.

\subsubsection{Assertion Monitors}\label{s:assertionMonitors}

For requirements checking, assertion monitors were implemented as state machines in Python, allowing sequences of events to be captured. If the precondition of an assertion is satisfied, the machine transitions to check the relevant postconditions, to determine whether the assertion holds or not. Otherwise, the postconditions are never checked.

For example, \textbf{Reqs.\ 1a-b} and \textbf{Req.\ 3} were both initially monitored as the following sequence:

\scriptsize
\begin{verbatim}
if (sensors_OK)
    wait_for(robot_decision)
    assert(robot_released_object)
\end{verbatim}
\normalsize
Note that, as with the logical properties, there may be different ways to implement an assertion for the same textual requirement, and there is scope for misinterpretation.

The results of the assertion checks, if triggered, are collected and a
conclusion about the satisfaction of the verified requirements can be drawn at
the end of simulation. 
The number of times each assertion monitor has been triggered in a set of tests can be used as a measure of the coverage achieved by that test set.

\subsection{Experiments}\label{s:exptDescription}

\bert{} can be verified experimentally with respect to the textual requirements
using a custom facility at the Bristol Robotics Laboratory, as shown in Figure~\ref{f:bert}.
When seeking to verify probabilistic properties of a system, the experiments should ideally provide an unbiased sampling representative of the system's deployed environment.
However, some phenomena may be difficult to reproduce naturally in experiments due to their rarity, safety considerations, or other practical limitations.
Consequently, experiment-based estimates of their likelihood may be inaccurate, as may estimates of dependent properties such as the overall success rate of the task.

\new{In the case of the handover task, we cannot confidently seek an overall success rate that accounts for the full possible range of conditions relating to hardware, software, the environment and the human (including mood, anatomy, and level of understanding of the task).
Human factors are particularly challenging to test in an unbiased way.
This problem can be ameliorated by acknowledging the constraints of the experiments or proactively constraining them to achieve a more reliable characterisation of a subset of the system's state space. 
The constraints become a part of the resulting  V\&V evidence. 
Thus, the experiments deliver an estimate of ``success rate within some set of constraints,'' instead of an estimate of ``overall success rate.'' 
More affordable or coverable V\&V tools such as simulation or formal modelling may be employed to gain confidence beyond this constrained experiment. Additionally, more detailed experiments may be performed to explore a wider range of human factors affecting the scenario, and to determine the overall success rate of the handover task beyond constraints. This is beyond the scope of this paper, however.}

As we were focusing on the ``typical use case'' of the handover scenario, in which the human has a working familiarity with the robot and intends to complete the task successfully, experiments were constrained accordingly.
\new{Each of the 10 subjects} was given clear instructions on how to successfully complete the task, followed by a practice session which ended when the task was successfully completed three times in a row.  
Subjects were instructed to try to complete the task successfully in each test.
All subjects confirmed that they had no physical disability that would affect their interaction with the robot.  
The robot started each test in a random pose.  
The object was placed in a fixed location, with random orientation about its vertical axis (thus changing the orientation of the optical markers, potentially affecting sensing of the object or influencing human hand placement on grasping).

Approval for experiments with volunteer subjects was obtained from the University of the West of England's Ethics Committee before they took place.  
A large, diverse cohort enables a more comprehensive V\&V to be carried out, but a cohort of 10 adult \new{volunteers} was deemed sufficient for the purpose of demonstrating \new{corroborative V\&V}.  
We recruited the volunteers from the Bristol Robotics Laboratory and the local area.
Most had prior robotics experience: three were post-graduate robotics students, one was a robotics entrepreneur, four were post-doctoral roboticists. Two had no prior experience of robotics.
All subjects signed a consent form prior to participation.

\subsubsection{Textual Requirements}\label{s:textualRequirements}

For physical experiments, \textbf{Reqs.\ 1--3} can be verified in their textual form based on visual observation, informed by video recordings and user feedback as necessary, e.g., to judge whether the human was ready or whether something has gone wrong.

\textbf{Reqs.\ 4--8} refer to software or physical parameters that cannot be reliably monitored by visual observation.
It is therefore appropriate to implement objective monitoring to inform judgements as to whether the textual requirements are satisfied.
To this end, ROS's built-in \texttt{rosbag} package was used to record all sensor readings, actuation signals, robot poses, and high-level control messages sent during each test.
Offline monitoring of these requirements was achieved by playing back the recordings while running assertion monitors as described in Section~\ref{s:assertionMonitors}.

In the case of \textbf{Reqs.\ 6--8}, these monitors depended on the robot's own sensing systems as the best available estimates of speed and spatial relationships.
In real-world V\&V exercises, independent sensing should be used.

\textbf{Req.\ 4} and \textbf{Req.\ 5} refer to the runtime behaviour of the robot's high-level control code.  
Hence the monitors used in simulation may also be applied to the experiment recordings, because the same robot code is used in each case.

All experiment recordings, along with the assertion monitor reports from simulations and experiments, are available from the University of Bristol's Research Data Repository~\citep{western2019simExptResults}.
 
\section{\new{Corroborative V\&V} of Requirements 1a and 1b}\label{s:verif}

After generating assets for V\&V through different V\&V techniques, we can generate \new{corroborative V\&V} evidence about the handover scenario, according to the plan described in Section~\ref{s:plan}.%

\begin{figure}[hbt]
\centering
\includegraphics[width=1.03\columnwidth]{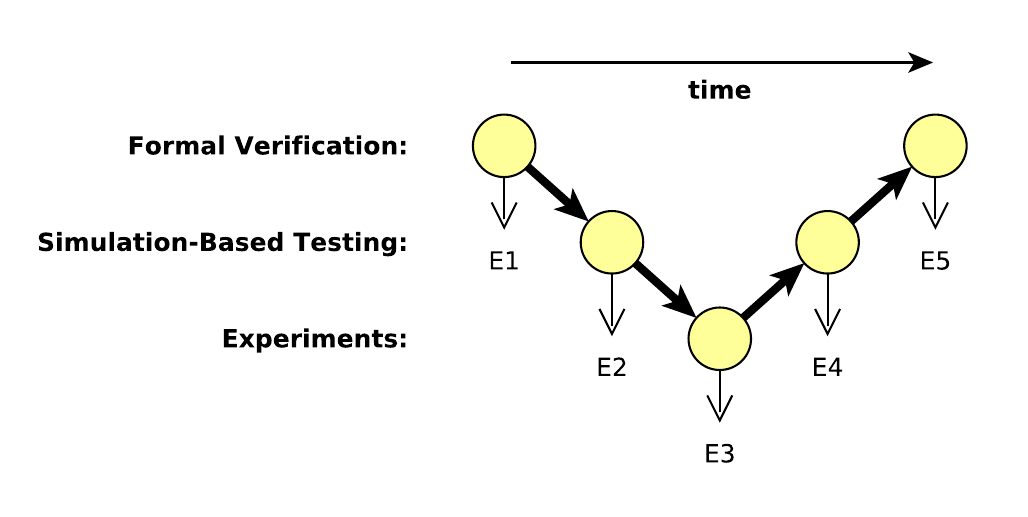}
\caption{A simplified representation of the \new{corroborative V\&V} workflow enacted in our case study, denoting the sequence in which evidence \A{1}--\A{5} were produced from individual V\&V techniques.}
\label{f:workflow}
\end{figure}

To discover whether the V\&V techniques \new{corroborate} one another, we compare evidence of the handover success rate (\textbf{Reqs.\ 1a-b}) from formal verification (\A{1}) and simulation-based testing (\A{2}).
Sources of discrepancy are identified and investigated in experiments with the physical system.
Experiment-based verification of the handover success rate in the `typical use case' (\A{3}) is then generated.
More detailed system characteristics measured during these experiments are used to inform modifications to the simulator, leading to  new evidence (\A{4}) that agrees closely with \A{3}.
These simulations also reveal a new aspect of the system's behaviour.
All insights gained up to this point are then used to inform modifications to the formal model, and the resulting evidence (\A{5}) is found to agree closely with \A{3} and \A{4}, satisfying our objective of achieving \new{corroboration} between the three V\&V techniques.
The enacted workflow, depicted in Figure~\ref{f:workflow}, is described in detail in the subsequent subsections.

\subsection{Formal Verification: Evidence \A{1}}
\label{s:fv1}

As described in Section~\ref{s:formalmodel}, the formal model includes probabilities of certain events coming to pass. 
Using the probabilistic model of the handover scenario we are able to determine that handover has close to 100\%
success rate:
\begin{equation}\label{e:fv1}
P (~%
\F \pred{handoverSuccessful} \\
~) = 0.9999948082592586
\end{equation}
That is, almost 100.0\% of the time the handover task completes successfully. This is a very high probability of success, meaning that there are very few paths through the model which result in failure of the handover task. There are two reasons for this. First, the model is based on a typical use case (see Section~\ref{s:plan}) in which the human's gaze, hand pressure and location are assumed to be  correct at all times. This reduces the likelihood of handover failure. Second, the robot waits for all of its sensors to report that gaze, pressure and location are correct before releasing its gripper. If any of these sensors does not report an acceptable value, then the robot continues to wait. This continues until the modelled robot eventually ``times out'' after 100 seconds. Given that the human always responds correctly in this version of the model, and there are no other sources of unreliability in the model, the only way the model can fail is if the robot times out while waiting for the sensors to report that the human's gaze, pressure and location are within acceptable bounds. As there are far many more paths through the model in which the handover completes successfully, the probability of success is very close to 100.0\%.

The formal model has shown that \bert{}   satisfies \textbf{Req.\ 1a} and \textbf{Req.\ 1b}:
\begin{itemize}[leftmargin=2cm]
\item [\R{1a}:] At least 95\% of handover attempts should be completed successfully.
\item [\R{1b}:] At least 60\% of handover attempts should be completed successfully.
\end{itemize}
However it is important to note that the formal model is using very rough estimates of the sensor reliabilities. To improve the accuracy of the formal model it is necessary to find more accurate figures for the sensor reliability. These could be obtained from manufacturer specifications, or through experiments with the \bert{} robot.

Despite the shortcomings of the formal model in its current form, we can still derive V\&V evidence, which we call \A{1}:
\begin{quote}
\label{s:assuranceFV1}
\A{1}: the success rate of handover is 100.0\%.
\end{quote}

\subsection{Simulation: Evidence \A{2} does not \new{corroborate} \A{1}}
\label{s:inaccuracies}

Evidence \A{1} can now be verified by another V\&V technique. In this case we use simulation as it is less costly than experimentation. 

Visual inspection of preliminary simulations indicated that the object sometimes fell from the robot's hand upon grasping or during carrying (``grip failure''), a possibility not previously considered.  
A new assertion monitor was constructed to capture this event in isolation.  
Additionally, the monitor for \textbf{Req. 1} was adapted to the form below to account for the possibility of grip failure.
Compared with the initial implementation presented in Section~\ref{s:simulationBasedTesting}, an earlier precondition is used to trigger the monitor: (\scriptsize\verb!robot_grasps_object!\normalsize).  
The original precondition (\scriptsize\verb!sensors_OK!\normalsize) is now asserted as a postcondition and is preceded by an additional postcondition (\scriptsize\verb!object_contacts_robot_hand!\normalsize) which is asserted repeatedly until sensing is complete.
This ensures that a verdict of `False' will be returned if the robot drops the object prematurely, regardless of any subsequent behaviour.

\scriptsize
\begin{verbatim}
if (robot_grasps_object)
    while !sensing_Done
        assert(object_contacts_robot_hand)
    assert(sensors_OK)
    wait_for(robot_decision)
    assert(robot_released_object)
\end{verbatim}
\normalsize
In a set of 100 simulations of the handover task, 80 attempts then were completed successfully. 
This result forms evidence \A{2}:
\begin{quote}
\A{2}: the success rate of handover is 80\%.
\end{quote}
Note that \A{1} and \A{2} disagree with each other, and are therefore \new{not corroborative}.
As explained in Section~\ref{s:workflows}, there are a number of potential causes of such a disagreement: inaccuracies in either the system models or the requirement models, or in the tools. The latter becomes more unlikely when established tools are used. 

In our case, the occurrence of grip failure was clearly the main source of discrepancy. A modelling inaccuracy was present in at least one of the two V\&V techniques used: the formal model implicitly assumed a grip failure rate of 0\%, whereas simulation indicated 20\%. \new{Both the formal model and the simulator assets were modified to account for this, as is shown in the following subsections.}

\subsection{Experiments: Evidence \A{3}}

\begin{newsection}
Before committing resources to user experiments, a set of hardware experiments was conducted to characterise the actual robot's grip failure rate.  
\bert{} was programmed to carry out the grasp-and-carry portion of the handover task 100 times, and grip failure was found to occur in 3 cases. 

At this point, despite some discrepancy, formal modelling and simulation were in agreement that the system satisfied the research-level minimum success rate of 60\%.
Furthermore, the simulation-based estimate was deemed likely to be conservatively low due to the inaccurately high grip failure rate.
It was therefore deemed worthwhile to proceed to user experiments.
\end{newsection}

User experiments were carried out as described in Section~\ref{s:exptDescription}. Results are summarized in Table~\ref{t:exptResults}.  To determine whether it was appropriate to treat our experimental data as independent samples of a single distribution, we investigated whether there was any noticeable effect of learning or prior robotics experience on the outcome of a test.
A statistical analysis was performed using IBM\textsuperscript{\textregistered} SPSS\textsuperscript{\textregistered} v23.0.
A Kruskal-Wallis H Test did not indicate a significant effect of the robotics experience categories on the number of handovers completed successfully after training ($\chi^2(3) = 1.5$, $p = 0.682$) or on the number of training runs required ($\chi^2(3) = 2.872$, $p = 0.412$).
Furthermore, a Spearman correlation revealed no significant correlation between the test number (1--10) and the total number of human-related failures in that test across subjects ($\rho = 0.220$, $p = 0.541$, two-tailed). 
It is possible that more extensive testing with a larger cohort would reveal weak but statistically significant effects of these parameters.
However, for the purposes of demonstrating our method, our cohort and experiment design was deemed to be adequate based on these results.

\begin{table*}[t]
\centering
\scriptsize
\renewcommand{\arraystretch}{1.3}
\caption{User experiment results \new{for the cohort of 10 volunteers}, by subject.
Robotics experience is denoted by N (none), S (post-graduate student), D (post-doctoral roboticist), or E (robotics entrepreneur).
Failure modes are denoted by R (robot grip failure), P (false negative pressure sensing), and L (false negative location sensing).}
\begin{tabular}{r|c|c|c|c|c|c|c|c|c|c|c}
Subject 	ID						& 1    & 2    & 3    & 4    & 5    & 6    & 7    & 8   & 9   & 10 & \textbf{mean} \\  \hline 
Experience 						& E    & D    & S    & N    & N    & D    & S    & D   & D   & S & \\ %
\# training runs 				& 5    & 4    & 6    & 6    & 6    & 3    & 3    & 6   & 3   & 6 & \textbf{4.8} \\  %
\# successes (post-training)		& 9    & 10   & 8    & 7    & 9    & 10   & 10   & 6   & 9   & 10 & \textbf{8.8} \\ %
Failure modes					& P    &     & P, L &R, P, P& R    &      &  & L, P, L, P & P & & \\
\end{tabular} 
\label{t:exptResults}
\end{table*}

The handover was successfully completed in 88 out of 100 tests.  
As in simulation, this can be taken as an estimate of the true success rate of the experimental system.

\begin{quote}
\A{3}: the success rate of handover is 88\% in the typical use case.
\end{quote}
Here the `typical use case' is that described in Section~\ref{s:exptDescription}.

Again we found notable disagreement between \A{3} and the previously generated evidence. 
A more specific discrepancy had already been identified in terms of the grip failure rate. 
To seek closer agreement between the three V\&V techniques, we explored the potential sources of discrepancy in greater detail. 

The video recordings and ROS logs, including sensor data, were reviewed to confirm the faults responsible for each failed handover in the user experiments.
The failure rate for each failure mode was identified as the number of occurrences divided by the number of opportunities for that fault to occur.
Results are listed in the first column of Table \ref{t:testingResults}.
`False negative' here was defined relative to the subject's observable actions.  
Thus false negative pressure sensing was identified where the review of logs and videos indicated that the subject was observably applying pressure to the object but the sensing threshold was not exceeded.  
Similarly, false negative location sensing was identified where the subject's hand was on the object during the sensing period but the robot's location sensor returned a negative result.
Rates of other possible failure modes (e.g., timeouts or false negative gaze sensing) are implicitly estimated to be 0\% based on these experiments.
This should not be taken as evidence that these modes never occur, only that they are rare.
Also, rates of false positive sensor readings could not be defined because, after training, there were no cases in which the subject did not apply their gaze, pressure, and hand location according to the protocol.

\subsection{Modifying the Simulator Asset}\label{s:modifyingsimulator}

The observed rates for individual failure modes were taken as the best available estimates of those properties in the typical use case and were used to tune the simulator asset (and, subsequently, the formal model asset) to represent that case.

In the previous simulations, the grip failure rate of 20\% was clearly much higher than the experimental observation of 3\%, while the simulated sensing did not reproduce the other observed failure modes.
Several aspects of the simulator were refined with the aim of approximating the experimentally observed rates of individual failure modes without sacrificing realism.

The accuracy of the simulated dynamics of the robot's handling of the object was improved by replacing default/placeholder values with more realistic estimates of inertial properties, material properties, and joint torque/velocity limits.

The instances of false negative 'location' sensing were identified as arising from the motion tracking system briefly losing track of the object (hand location is measured relative to the object) and reassigning its location to another point.  Mimicking this behaviour, the simulated motion tracking was set to reassign the observed location of the object (but not the person's hand or head) to an arbitrary point in 3.1\% of readings.  

Based on the recordings, all cases of false negative pressure sensing seen in the user experiments were attributed to the subject pulling on the object more gently than in other cases.
The exact forcing pattern applied by the subjects could not be extracted from the experiment data. 
Instead, the lower threshold of the distribution from which the simulated human pulling force was selected was lowered from 5 N to 1 N through a process of trial-and-error to approximate the failure rate seen in user experiments.

After tuning, a set of 500 simulations was run.
In all tests the simulated human enacted the trace of high-level actions corresponding to the typical use case, remaining engaged in the task and applying their gaze, pressure, and location within the relevant bounds.
The results, included in Table \ref{t:testingResults}, indicate that the tuning process was successful in approximating the individual failure rates observed in user experiments.
Close \new{corroboration} is also achieved in the handover success rate, although it must be acknowledged that this correspondence slightly overestimates the true accuracy of the simulator;
larger errors are seen in the rates of individual failure modes.
Nevertheless, we have improved confidence in the simulation as a representation of the physical system and in the \new{corroborative evidence} provided by each V\&V technique.
\A{3} is now supported by new evidence from simulation-based testing:

\begin{quote}
\A{4}: The success rate of handover is 87.8\% in the typical use case.
\end{quote}
\begin{table}[ht]
\centering
\scriptsize
\renewcommand{\arraystretch}{1.3}
\caption{Test outcomes and occurence rates of individual failure modes for the typical use case, according to 100 user experiments and 500 simulations after tuning.}
\begin{newsection}
\begin{tabular}{r|c|c}
								& User experiments	& Simulation			\\  \hline 
Number of tests					& 100				& 500 				\\  %
Handover success					& 88.0\% (88/100)	& 87.8\% (439/500)	\\  %
Runtime error					& 0.0\% (0/100)		& 0.2\% (1/500)		\\  %
Grip failure		 				& 2.0\% (2/100)		& 1.6\% (8/499)		\\  %
False negative gaze				& 0.0\% (0/98)		& 0.0\% (0/491)		\\  %
False negative pressure			& 7.1\% (7/98)		& 6.5\% (32/491)		\\  %
False negative location			& 3.1\% (3/98)		& 4.2\% (21/491)		\\  %
\end{tabular} 
\end{newsection}
\label{t:testingResults}
\end{table}
Furthermore, the simulations exposed a failure mode not previously considered.
In one test, the handover success monitor returned no result and inspection of the logs revealed that the robot's control code crashed due to a runtime error:
\scriptsize
\begin{verbatim}
RuntimeError: Unable to connect to move_group action 
    server 'place' within allotted time (2)
\end{verbatim}
\normalsize
This message indicates that a timeout occurred when invoking the robot's motion planning module.
The robot's high-level control code does not include any means of handling such exceptions.
Although rare, these events may significantly affect the user's trust if they occur in deployment, and could lead to violations of critical safety requirements.
In our case, the error caused the only violations of \textbf{Reqs.\ 4--5}.
The exposure of the error, which required high-volume testing and a realistic implementation of the system, demonstrates a key strength of simulation as a complement to formal modelling and user experiments.
It is conceivable that the error never occurs in the actual system, e.g., due to differences in computational load during simulation.
However, further testing on the real system cannot rule out the possibility completely.
A more conservative approach is to adopt the simulation-based estimate of the error's frequency as the basis for further \new{corroborative V\&V} and design recommendations.

\subsection{Modifying the Formal Model Asset}

Now that we have verified determined simulation evidence \A{3}, we can attempt to \new{corroborate it} using formal verification to address the discrepancy discovered between \A{1} and \A{3} during the first V\&V cycle. As described in Section~\ref{s:fv1}, evidence \A{1} generated by formal verification disagrees with evidence \A{3}, generated by experiments:
\begin{quote}
\A{1}: the success rate of handover is 100.0\%.

\A{3}: the success rate of handover is 88\% in the typical use case.
\end{quote}
The formal model currently uses placeholder estimates for the reliability of the gaze, pressure and location sensors on the \bert{} robot. However, using some of the experimental data in Table~\ref{t:testingResults}  it is possible
to replace the corresponding estimates in the formal model with more
accurate values. In particular, we can use the following values:

\begin{itemize}
\item Gaze sensor, false negative: 0.0\%
\item Pressure sensor, false negative: 7.1\%
\item Location sensor, false negative: 3.1\%
\end{itemize}
False negatives and true positives are mutually exclusive, since the former refers to when the person's gaze/pressure/location is correct but the sensor reports (incorrectly) that it is not, and the latter refers to when the person's gaze/pressure/location is correct and the sensor reports (correctly) that it is. Therefore, we can infer true positive values:

\begin{itemize}
\item Gaze sensor, false negative: 0.0\%, true positive: 100.0\%
\item Pressure sensor, false negative: 7.1\%, true positive: 92.9\%
\item Location sensor, false negative: 3.1\%, true positive: 96.9\%
\end{itemize}
As the experiments did not report any situations where there were false positives, we assume that the rate of false positive sensor failures is 0.0\% for each sensor, and therefore the rate of true negatives for each sensor is 100.0\%.

We can now set the probabilities in the model accordingly:

\begin{scriptsize}
\begin{Verbatim}[tabsize=4]
const double pGazeFN			= 0.00;			
const double pGazeTP			= 1-pGazeFN;
const double pGazeFP			= 0.00;   		
const double pGazeTN			= 1-pGazeFP;
const double pPressureFN		= 0.071428571;   		
const double pPressureTP		= 1-pPressureFN;
const double pPressureFP		= 0.00;   	
const double pPressureTN		= 1-pPressureFP;
const double pLocationFN		= 0.030612245;   		
const double pLocationTP		= 1-pLocationFN;
const double pLocationFP		= 0.00;		
const double pLocationTN		= 1-pLocationFP;  
\end{Verbatim}
\end{scriptsize}
Verifying the model we can obtain the success rate of the handover task:
\begin{equation}\label{e:fv2}
P (~\F \pred{handoverSuccessful}~) = 0.9999954384256133
\end{equation}
It can be seen that the success rate remains at almost 100.0\%. This is to be expected as the sensor failure rates have changed slightly, but it remains the case that the only way for the handover to fail is for the robot to time-out. 

There is still a significant difference between this success rate and the success rate reported by simulation (87.8\%) and experiments (88\%). This may be, in part, due to the way in which the sensors were modelled in the formal model. It was assumed that sensors might make any number of ``samples'' while the robot is waiting for the person to grasp the object in the correct way. Each one of these samples is a separate event, in which the sensor takes a reading which is reported back to the robot's decision-making system. Therefore, each time the sensor takes a reading there is a probability of failure, and false positives and negatives are possible. The formal model reflects this, and the failure rates above apply to each reading taken by the sensor, rather than the average failure rate per handover. The \prism{} code defining the gaze sensor module was as follows:

\begin{scriptsize}
\begin{Verbatim}[tabsize=4]
module gazeSensor
	gazeSensorState : [0..1000] init null;
	
	[senseGaze] robotState=waitForGPLUpdate &
	  gazeState=gazeOk -> pGazeFN: (gazeSensorState'=
	    gazeNotOk) + pGazeTP: (gazeSensorState'=gazeOk);

	[senseGaze] robotState=waitForGPLUpdate &
	  gazeState=gazeNotOk -> pGazeTN: (gazeSensorState'=
	    gazeNotOk) + pGazeFP: (gazeSensorState'=gazeOk);
endmodule
\end{Verbatim}
\end{scriptsize}
The first transition rule says that if the robot is currently waiting for the person to grasp the object (\texttts{waitForGPLUpdate}) and the gaze is okay (i.e., the person is looking in the right direction), then update the value of \texttts{gazeSensorState} to either \texttts{gazeOk} or \texttts{gazeNotOk}, depending on the probability of false negative and true positive. The second transition rule does something similar for the case where the person is not looking in the right direction. Note that the only guards on these transitions specify that the robot is waiting for the person to grasp the object and that the gaze is either okay or not okay. (The synchronisation \texttts{senseGaze} is simply used to keep track of how long sensing is taking within a timekeeper module and is not relevant in this example.) Therefore, these sensor readings can happen any number of times while the robot is waiting for the person to be ready to receive the object and complete the handover task.

This way of modelling the handover scenario produces less accurate results when combined with the failure rates established by experiment. This is because the failure rates determined were based on the number of experiments in which, for example, the pressure sensor was seen to give a false negative reading. For example, the probability of 0.071 for a pressure sensor false negative reading was obtained by dividing the number of experiments in which a false negative reading occurred {at some point} (7) by the total number of experiments not interrupted by gripper failure (98).

Therefore it would be more accurate to re-model the scenario in a way that reflects experimental reality; that is, the probability of a sensor failure for a handover of the object should be based on the observed average rate of failure of that sensor. This was achieved by modifying the gaze, pressure and location sensor modules in the \prism{} model:

\begin{scriptsize}
\begin{Verbatim}[tabsize=4]
module gazeSensor
	gazeSensorState : [0..1000] init null;
	
	gazeSensorSet: bool init false;
	[senseGaze] robotState=waitForGPLUpdate &
	  gazeState=gazeOk & !gazeSensorSet ->
	  pGazeFN: (gazeSensorState'=gazeNotOk) &
	  (gazeSensorSet'=true) + pGazeTP:
	  (gazeSensorState'=gazeOk) & (gazeSensorSet'=true);
	  
	[senseGaze] robotState=waitForGPLUpdate &
	  gazeState=gazeNotOk & !gazeSensorSet ->
	  pGazeTN: (gazeSensorState'=gazeNotOk) &
	  (gazeSensorSet'=true) + pGazeFP:
	  (gazeSensorState'=gazeOk) & (gazeSensorSet'=true);
endmodule 
\end{Verbatim}
\end{scriptsize}
In this revised model each sensor's state can be set only once. For example, for the gaze sensor, this is done by introducing a Boolean variable \texttts{gazeSensorSet} which is initally false, but is set to true once a sensor reading has been taken, and is never again set to false. Therefore, this model reflects the experiments more closely.

Verifying the new model gives us a new value for the reliability of the handover task:
\begin{equation}\label{e:fv3}
P (~\F \pred{handoverSuccessful}~) = 0.9001457729154516
\end{equation}
The handover task now completes successfully with a probability of 90.0\%. This is closer to the simulation and experiment results of 87.8\% and 88.0\% respectively, but there is still a noticeable difference. One possible reason for this is that the gripper failure rate, as determined by experiment and built into the simulation, is not yet modelled in \prism{}. The following transition describes what happens within the robot module once the gaze, pressure and location are found to be correct:

\begin{scriptsize}
\begin{Verbatim}[tabsize=4]
[tick] robotState=GPLOk -> (handContents'=nothing) &
  (robotState'=handoverSuccessful);
\end{Verbatim}
\end{scriptsize}
Here, once the robot's state reaches \texttts{GPLOk}, indicating that gaze, pressure and location are within acceptable bounds, the robot releases its gripper and hands over the object to the person. Therefore the \texttts{handContents} variable is updated to reflect that the robot's hand/gripper is now empty, and the robot's state is updated to show that handover has been successful. To introduce the possibility of gripper failure, this transition was modified to incorporate a probabilistic choice:

\begin{scriptsize}
\begin{Verbatim}[tabsize=4]
[tick] robotState=GPLOk -> pGripperOk: (handContents'=
  nothing) & (robotState'=handoverSuccessful) +
  pGripperFailure: (handContents'=nothing) &
  (robotState'=handoverUnsuccessful);
\end{Verbatim}
\end{scriptsize}
Now, one of two things can happen. The first possibility is that the handover completes successfully, as before, with a probability of \texttts{pGripperOk}. The second is that the handover fails, with probability \texttts{pGripperFailure}. These two probabilities are set like so:

\begin{scriptsize}
\begin{Verbatim}[tabsize=4]
const double pGripperFailure = 0.02;
const double pGripperOk = 1 - pGripperFailure;
\end{Verbatim}
\end{scriptsize}
Here, ``pGripperFailure'' is set to 0.02 in accordance with the gripper failure rate of 2\% determined by experiment (see Table~\ref{t:testingResults}). We verify the model once again to determine a handover success rate of 88.2\%:
\begin{equation}\label{e:fv4}
P (~\F \pred{handoverSuccessful}~) = 0.8821428574571426
\end{equation}
In a similar way a new transition was introduced into the transition system to model the possibility of failure of \bert{}'s motion planning module, as described in Section~\ref{s:modifyingsimulator}. This transition occurs at the start of the handover task as the robot prepares to move its arm to grasp the object for handover. The revised transition rule incorporates probabilities for the success or failure of the motion planning module:

\begin{scriptsize}
\begin{Verbatim}[tabsize=4]
[activateRobot] robotState=waiting -> pMotionOk:
  (robotState'= moveHandToObjectLocation) +
  pMotionFailure: (robotState'=motionError);
\end{Verbatim}
\end{scriptsize}
These probabilities were based on the data shown in Table~\ref{t:testingResults}:

\begin{scriptsize}
\begin{Verbatim}[tabsize=4]
const double pMotionFailure = 0.002;  // 0.2%
const double pMotionOk = 1 - pMotionFailure;
\end{Verbatim}
\end{scriptsize}
Verifying the model once more gives an updated handover success rate of 88.0\%:
\begin{equation}\label{e:fv5}
P (~\F \pred{handoverSuccessful}~) = 0.8803785717422283
\end{equation}
Thus the final evidence provided by formal verification may be stated as:

\begin{quote}
\A{5}: the success rate of handover is 88.0\% in the typical use case.
\end{quote}
After conducting \new{corroborative V\&V} of the handover task for the \bert{} system, it was found that all V\&V techniques were \new{corroborative} on the probability of a successful handover. The probabilities are shown in Table~\ref{t:results}.

\begin{table}[ht]
\centering
\caption{Results of \new{corroborative V\&V}.}
\label{t:results}
\begin{tabular}{l|l}
Formal Verification & 88.0\% \\
Simulation & 87.8\% \\
Experiments & 88.0\% \\
\hline
Average & 87.9\% $\pm$ 0.1\%
\end{tabular}
\end{table}

\noindent
Having established confidence in our models using \new{corroborative V\&V}, we can assert that in the typical use case, \textbf{Req. 1b} is satisfied but \textbf{Req. 1a} is not.

\section{V\&V of Requirements 2--8}
\label{s:verification-other-reqs}

In the previous section we focused our efforts  on the V\&V of \textbf{Reqs. 1a--b} in order to demonstrate \new{corroborative V\&V} of a robotic system. However, for the sake of completeness and in line with  best practices in engineering, we also attempted V\&V of \textbf{Reqs. 2--8} using each of the three V\&V techniques. These V\&V results are presented without reference to \new{corroboration, but corroborative V\&V} could be applied to \textbf{Reqs. 2--8} in a similar manner to \textbf{Reqs. 1a--b}.

It can be seen in the following sub-sections that the different V\&V techniques do not all agree on how well \textbf{Reqs. 2--8} are met. This is similar to the case study for \textbf{Reqs. 1a--b} before \new{corroborative V\&V}. Given enough time, it would be possible to apply the \new{corroborative V\&V} approach to this expanded set of requirements in order to find the source of the disagreements between V\&V techniques and to improve the \new{level of corroboration} between them.

\subsection{Experiments}

For the user experiments, the full set of textual requirements was evaluated through a combination of offline assertion monitoring and visual observation, as described in Section~\ref{s:textualRequirements}.
Table~\ref{t:expAssnResults1} presents the verdicts returned from each individual test.
\textbf{Req.\ 1} is included for completeness.
Note that for \textbf{Reqs.\ 4--8}, up to seven of the missing verdicts were attributable to errors in the recording process rather than the tests themselves.

\begin{table}[ht]
\centering
\scriptsize
\renewcommand{\arraystretch}{1.3}
\caption{User experiments: Results on textual requirements from 100 tests.
``Covered'' indicates the number of tests from which a verdict could be achieved.
``Passed'' and ``Failed'' indicate the number of tests in which the requirement was deemed to be satisfied or violated, respectively.
``Pass rate'' is calculated as the ratio ``Passed'':``Covered''.}
\begin{tabular}{c|c|c|c|c|c|c|c|c}
\textbf{Req.} & \textbf{1}   & \textbf{2}  & \textbf{3}  & \textbf{4}  & \textbf{5}  & \textbf{6}  & \textbf{7}  & \textbf{8}  \\
\hline %
Covered	  & 100  & 0  & 98 & 93 & 93 & 25 & 98 & 90 \\
Passed	  & 88   & 0  & 88 & 93 & 93 & 25 & 78 & 90 \\
Failed	  & 12   & 0  & 10 & 0  & 0  & 0  & 20 & 0  \\
Pass rate & 0.88 & --- & 0.9 & 1.0 & 1.0 & 1.0& 0.8& 1.0 \\
\end{tabular}
\label{t:expAssnResults1}
\end{table}

\noindent
As noted previously, the handover success rate in the user experiments satisfies \textbf{Req.\ 1b} but violates  \textbf{Req.\ 1a}.
Correspondingly, violations of \textbf{Req. 3} arise from the cases of false negative sensor readings.
Additionally, we see that \textbf{Req. 7} is violated in 78 out of 98 tests;
the robot occasionally violates its speed threshold upon resetting, presumably depending on its initial pose.
A notable ``coverage hole'' is seen in this test set for \textbf{Req. 2}, as the human was judged to be ready for the handover in every test.
All other requirements were covered in at least 25 tests, and no other violations were observed.

\subsection{Simulation-based Testing}

Table~\ref{t:simAssnResults1} presents the results of the assertions monitored in the same 500 simulation-based tests summarised in Table~\ref{t:testingResults}, representing the typical use case. Comparing Table~\ref{t:simAssnResults1} with the experiment results in Table~\ref{t:expAssnResults1}, we see broad \new{corroboration}, but with several noteworthy discrepancies discussed below.

\begin{table}[ht]
\centering
\scriptsize
\renewcommand{\arraystretch}{1.3}
\caption{Simulation: assertion coverage and results corresponding to each of the requirements (corrected for missing results) in a set of 500 tests.}
\begin{tabular}{c|c|c|c|c|c|c|c|c}
\textbf{Req.} & \textbf{1}   & \textbf{2}  & \textbf{3}  & \textbf{4}  & \textbf{5}  & \textbf{6}  & \textbf{7}  & \textbf{8}  \\
\hline %
Covered	& 500  & 53    & 446 & 500 & 500 & 500 & 500 & 0 \\
Passed	& 439  & 53    & 446 & 499 & 499 & 500 & 437 & 0 \\
Failed	& 61   & 0     & 0   & 1   & 1   & 0   & 63  & 0 \\
Pass rate & 0.878 & 1.0& 1.0 & 0.998 & 0.998 & 1.0 & 0.874 & --- \\ 
\end{tabular}
\label{t:simAssnResults1}
\end{table}

All assertions were covered --- i.e., all monitors were triggered at least once --- except for \textbf{Req.\ 8}.
This indicates that the human and robot did not come within 10 cm of each other during the interaction.
While this is possible given the length of the object to be handed over, the experiments revealed that closer proximities are seen in typical use.
Hence this constitutes a notable coverage hole in these tests.

Contrary to the experiment results,  \textbf{Req.\ 2} was covered in several tests and no violations of  \textbf{Req.\ 3} were observed.
Further investigation of this discrepancy revealed a potential requirements inaccuracy;
the assertion corresponding to this requirement expressed ``the human is ready'' as \verb!sensors_ok!.
In this sense, the assertion monitor verifies only the high-level control of the robot, discounting the possibility of sensor errors.
Hence the results are still informative, but some modification of the assertion monitors would be required to achieve a more comprehensive V\&V of these requirements.

As noted previously,  \textbf{Req.\ 4--5} were violated by the single runtime error.
The observation that  \textbf{Req.\ 7} is violated in 63 out of 500 tests is consistent with the experiment results.

\subsection{Formal Verification}

\textbf{Req.\ 2} says that if the human is not ready, the robot shall not hand over the object. It was formalised as follows:

\begin{scriptsize}
\begin{equation}
\label{e:fv:req2}
P\left(\G\left[
\begin{array}{c}
\neg \left(
\begin{array}{c}
\pred{gazeState}=\pred{gazeOk}~\wedge\\
\pred{pressureState}=\pred{pressureOk}~\wedge \\
\pred{locationState}=\pred{locationOk}
\end{array}\right)\\
	\Longrightarrow\\
	\begin{array}{c}
	\neg\left(
	\begin{array}{c}
	\pred{robotState}=\pred{handoverSuccessful}~\vee \\
	\pred{robotState}=\pred{handoverUnsuccessful}
	\end{array}\right)
	\end{array}
\end{array}
\right]\right)
\end{equation}
\end{scriptsize}%
This property says that it is always the case that if the human's gaze, pressure and hand location are not correct, then it is not the case that the robot has attempted to hand over the object. (Handing over the object results in either the ``handoverSuccessful'' or ``handoverUnsuccessful'' states.)
Verifying this property in \prism{} gives a probability of 1.0, meaning that it is always true.

\textbf{Req.\ 3}, which says that, ``if the human is ready, the robot shall hand over the object,'' was formalised in a similar way:

\begin{scriptsize}
\begin{equation}
\label{e:fv:req3}
P\left(\G\left[
\begin{array}{c}
\left(
\begin{array}{c}
\pred{gazeState}=\pred{gazeOk}~\wedge\\
\pred{pressureState}=\pred{pressureOk}~\wedge \\
\pred{locationState}=\pred{locationOk}
\end{array}\right)\\
	\Longrightarrow\\
	\begin{array}{c}
	\F\left(
	\begin{array}{c}
	\pred{robotState}=\pred{handoverSuccessful}
	\end{array}\right)
	\end{array}
\end{array}
\right]\right)
\end{equation}
\end{scriptsize}%
It was expected that this property would be evaluated by \prism{} as less than 1.0 due to the possibility of sensor and gripper failures. Indeed, verification using \prism{} gave a result of 0.8803785717422283.

\textbf{Req.\ 4} states that the robot always reaches a decision within a threshold of time. This was formalised as follows:

\begin{scriptsize}
\begin{equation}
\label{e:fv:req4}
P\left(\G\left[
\begin{array}{c}
\left(
\begin{array}{c}
\pred{robotState}=\pred{GPLOk}
\end{array}\right)\\
	\Longrightarrow\\
	\begin{array}{c}
	\F\left(
	\begin{array}{c}\left(
	\begin{array}{c}
	\pred{robotState}=\pred{handoverSuccessful}~\vee\\
	\pred{robotState}=\pred{handoverUnsuccessful}
	\end{array}\right)\\
	\wedge~\pred{robotState}=\pred{objectReleaseTimer} \le 20\\
	\end{array}\right)
	\end{array}
\end{array}
\right]\right)
\end{equation}
\end{scriptsize}%
Here the phrase ``reaches a decision'' was taken to mean that the robot has decided to release the object. In the model, this can result in ``handoverSuccessful'' if the gripper works properly or ``handoverUnsuccessful'' if the gripper fails. The requirement specifies that this should happen within ``a threshold of time'' but does not specify the amount of time. In our model, we specified that the gripper release would take 2.0 seconds based on consultation with the robot's users. Time was quantified in the model using a ``objectReleaseTimer'' which is set to zero when the robot determines that the humans' gaze, pressure and location are acceptable. The objectReleaseTimer was set to work in 0.1 second intervals in order to provide adequate precision without increasing the size of the state space to intractable levels. Therefore the property above captures \textbf{Req.\ 4} as it states that once the robot has found the human's gaze, pressure and location to be acceptable, then it will attempt to release the gripper (either successfully or unsuccessfully) within 2.0 seconds.

This property was verified and the probability was determined to be 0.9999999999999996, or 100.0\% allowing for floating point arithmetic precision errors in \prism{}'s computation engine~\citep{prism-manual}.

\textbf{Req.\ 5} states that the robot shall always either time-out, decide to release the object, or decide not to release the object. It was formalised as follows:

\begin{newsection}
\begin{scriptsize}
\begin{equation}
\label{e:fv:req5}
P\left(\G\left[
\begin{array}{c}
(\F\pred{robotState}=\pred{handoverSuccessful})~\vee\\
(\F\pred{robotState}=\pred{handoverUnsuccessful})~\vee\\
(\F\pred{robotState}=\pred{waitForGPLUpdate} \U\\
\pred{robotState}=\pred{timedOut})
\end{array}
\right]\right)
\end{equation}
\end{scriptsize}%
\end{newsection}%
This property specifies the probability that it is always the case that the robot eventually decides to release the object (either successfully or unsuccessfully) or times-out while waiting for the human's gaze, pressure and location to update to acceptable values.  The latter case, where the robot times out, is effectively the same as the robot deciding to not hand over the object.

This property was verified revealing a probability of 0.9979999999999996, which was expected \new{as the runtime error  encountered in Section~\ref{s:modifyingsimulator}, which was also included in the PRISM model, has a failure rate of 0.2\% or 0.002.} In order to check that this was the case, another property was specified \new{which says that the robot can behave as expected in the previous property, or eventually encounter a runtime error:}

\begin{scriptsize}
\begin{equation}
\label{e:fv:req5b}
P\left(\G\left[
\begin{array}{c}
(\F\pred{robotState}=\pred{handoverSuccessful})~\vee\\
(\F\pred{robotState}=\pred{handoverUnsuccessful})~\vee\\
(\F(\pred{robotState}=\pred{waitForGPLUpdate}~\U\\
\pred{robotState}=\pred{timedOut}))~\vee\\
(\F\pred{robotState}=\pred{runtimeError})
\end{array}
\right]\right)
\end{equation}
\end{scriptsize}%
This property was verified resulting in a probability of 0.9999999999999993, or 100.0\% allowing for precision errors.

\textbf{Req.\ 6} states that the robot shall not close its hand when the human is too close, \textbf{Req.\ 7} says that the robot shall start in a restricted speed mode, and \textbf{Req.\ 8} says that if the robot is within 10 cm of the human the robot's hand speed is less than 250 mm/s. 
\new{These properties could not be modelled, specified or verified formally as the PRISM model of the handover scenario does not include a model of a proximity sensor, and does not allow for speeds or distances to to be set within the control system.}
\new{%
It is possible, in principle, to re-model the scenario to include such detail. However, adding complexity to the model adds to the computational resources required to verify the model. In some cases, formal verification can become intractable. Therefore, it may be more practical for V\&V of  \textbf{Reqs.\ 6--8} to rely more heavily on evidence gained from simulation and experiment where physical properties can be much more fine-grained.}

\subsection{Computational Demands}

Properties \ref{e:fv1} to \ref{e:fv:req5b} were verified against several different \prism{} models representing the handover task. \new{These models were created during the corroborative V\&V process shown in Section~\ref{s:verif}.} The complexity of the \prism{} model checking for these properties is shown in Table~\ref{t:fv-complexity}. From left to right, the columns show the requirement and property verified, numbers of states and transitions used, time required for building the model and time/memory required to verify the model. %
PRISM 4.2.1 was used on an eight-core Intel\textsuperscript{\textregistered}
  Core\textsuperscript{TM} i7 laptop with 16 GB
  of memory running Ubuntu Linux 12.04.

It can be seen that Properties~\ref{e:fv:req3}--\ref{e:fv:req5b} took significantly longer to verify than the other properties. This is most likely the result of the use of nested temporal logic operators (e.g., $\F$, $\G$, $\X$, $\textsf{U}$) in these properties compared to Properties~\ref{e:fv1}--\ref{e:fv5}, which use simpler formulae. Properties~\ref{e:fv:req2}--\ref{e:fv:req5b} took the same time to build the model (69.4 seconds) as these properties were all checked against a single \prism{} model file, which needed to be built only once before these properties could be verified. The amount of memory used for Property~\ref{e:fv:req2} was not returned by \prism{}, so this value has been omitted from the table.

\begin{table}[ht]
\begin{newsection}
\centering
\scriptsize
\caption{\new{Complexity of formal verification using \prism{}.}}
\label{t:fv-complexity}
\begin{tabular}{c|c|r|r|r|r|r|r|}
\multicolumn{4}{c}{~} & \multicolumn{1}{|c|}{Build} & \multicolumn{2}{|c|}{Verification} \\\hline
Req. & Prop. & States & Transitions & T (s) & T (s) & M (kB) \\
\hline
1 & \ref{e:fv1} & 42,960 & 236,643& 36.8 & 0.203 & 2,253  \\
1 & \ref{e:fv2} & 31,120 & 150,955 & 61.4 & 0.147 & 1,741  \\
1 & \ref{e:fv3} & 15,614 & 54,969 & 84.0 & 0.062 & 997 \\
1 & \ref{e:fv4} & 15,615 & 54,971 & 90.0 & 0.057 & 999 \\
1 & \ref{e:fv5} & 15,623 & 54,998 & 40.6 & 0.053 & 1,003 \\
2 & \ref{e:fv:req2} & 15,623 & 54,998 & 69.4 & 0.002 & * \\   %
3 & \ref{e:fv:req3} & 15,623 & 54,998 & 69.4 & 20.3 & 1,536 \\
4 & \ref{e:fv:req4} & 15,623 & 54,998 & 69.4 & 17.7 & 1,007 \\
5 & \ref{e:fv:req5} & 15,623 & 54,998 & 69.4 & 24.9 & 1,843 \\
5 & \ref{e:fv:req5b} & 15,623 & 54,998 & 69.4 & 50.2 & 2,048 \\
\end{tabular}
\\** = Value not returned by \prism{}.
\end{newsection}
\end{table}%

\noindent
Simulation-based testing was performed using ROS Indigo and Gazebo v2.2.3 on a quad-core Intel\textsuperscript{\textregistered} Core\textsuperscript{TM} i7 laptop with 8 GB of memory running Ubuntu Linux 14.04.
With all online monitors running, simulations were executed at a speed of 0.8$\times$real-time on average, taking 69.3 seconds per test.
\new{Of course, with the advantages of batch and parallel processing,} simulation-based testing remains considerably faster than physical experiments.

\section{Discussion}\label{s:discussion}

\new{Through the \new{corroborative combination of a number of V\&V techniques}, namely formal verification, simulation-based testing, and experiments, we have determined the handover success rate (\textbf{Reqs.\ 1a--b}) with greater confidence than could be achieved by any of the V\&V techniques in isolation. Each of the different V\&V techniques was used iteratively to \new{corroborate} the evidence found by the other techniques during the \new{corroborative} V\&V process.
Although the experiments alone would have returned a similar value for the handover success rate, achieving \new{corroboration} in model checking and in simulation gives a higher level of confidence that the experimental results are correct and that the robot system meets its requirements.}

\new{The corroborative V\&V process exposed key differences between the models used in the V\&V techniques, specifically the false negative and true positive rates for the gaze, pressure and location sensors, as well as the grip failure rate.  
For \textbf{Reqs.\ 4--6}, the combination of simulation-based testing and formal verification exposed important system behaviours not observed in the experiments, i.e.,\ requirement violations. 
The observed runtime error (which caused violations of  \textbf{Reqs.\ 4--5}) could only be exposed through a high number of tests in simulation.
The subsequent inclusion of this error in the formal model and the simulator ensured that its impact on the behaviour of the system could be explored with more coverability using formal verification and simulation-based testing. Furthermore, corrected models for these two V\&V techniques were obtained to balance coverability capabilities, expressivity and realism.}

\new{Corroborative V\&V has demonstrated} that (i) the system satisfies \textbf{Req.\ 1b}, and (ii) the more stringent version, \textbf{Req.\ 1a}, is not satisfied, to a greater degree than if the individual V\&V techniques were used without \new{corroboration}.
Based on the insights gained during the V\&V process, several design recommendations could be made to improve the handover success rate and to satisfy other requirements.
The sensing process could be made more robust to sudden changes in the human motion, or to reduce the number of handover failures due to sensing errors through mechanisms such as ``debouncing'' for the sensor readings. \new{(Debouncing prevents a single event from creating multiple sensor signals.)} 
Adjustments to the robot's hardware or motion planning strategy may improve the gripper failure rate.
A speed limit needs to be introduced when the robot is reset, to avoid dangerous unintended collisions.
Also, as uncontrollable faults can be encountered during execution, we could instrument our code to perform diagnostics and fault recovery strategies.

For demonstration purposes, we focused on achieving \new{corroboration}  relating to a particular set of requirements, \textbf{Reqs. 1a--b}.
As our examination of \textbf{Reqs.\ 2--8} demonstrates, \new{corroboration} on some requirements does not entail \new{corroboration} across all requirements. 
For \textbf{Reqs. 1a--b}, the end result was that all V\&V techniques agreed on the success rate of handover within a range of $\pm 0.1\%$. In an ideal world all V\&V techniques use accurate models of the world, and are accurate with respect to one another, so that V\&V evidence generated with one technique should also be found valid by another. In practice, this may not happen. If two V\&V techniques do not agree, then we might look for inaccuracies in the system models, the requirements models, or the tools as described in Section~\ref{s:workflows}. However, after a number of iterations through the \new{corroborative V\&V} diagram (Figure~\ref{f:approach}), we may still have V\&V techniques in disagreement. One possible reason might be project constraints: we may lack the resources to continue to address inaccuracies. Another reason could be that the V\&V techniques may be lacking: for example, model checking for formal verification can often be hindered by the state space explosion which limits the accuracy of models that can be checked. Alternatively, we might lack the computational resources to explore sufficient numbers of simulated experiments, or we may lack the personnel to conduct sufficient numbers of real experiments.

Therefore, in practice, \new{corroboration between} V\&V techniques may not be possible. At this point we might assess whether our V\&V techniques are up to the job. Perhaps we should use an automated theorem prover rather than a model checker? Or perhaps a two-dimensional physical simulation would work better than a three-dimensional one? Perhaps we could create the simulation using a different programming language or use a more powerful computer? The list goes on.

We might also decide that exact \new{corroboration} is not necessary if all the V\&V techniques are within an acceptable range. For example, we might have three different pieces of evidence, each generated by a different V\&V technique:
\begin{quote}
 \A{$i$}: System reliability is 92\%.\\
 \A{$j$}: System reliability is 98\%.\\
 \A{$k$}: System reliability is 93\%.
\end{quote}
Clearly all three pieces of evidence are not in agreement with the others. However, the lowest value for system reliability given is 92\%, which means that all three V\&V techniques agree on the following statement:
 ``System reliability is 92\% or greater.''
Note that this statement is implicit in evidence \A{$i$}, \A{$j$} and \A{$k$}. In this case we have used \new{corroborative V\&V} to allow us to determine a minimum value for reliability. This value can then be checked with respect to system requirements to see whether the system being modelled is sufficiently reliable.

There may be other reasons, beyond those discussed in this section, why we cannot reach \new{corroboration} between V\&V techniques, and there may be other ways to remedy this beyond range-based statements like the one above. It is intended that the suggestions given here may provide direction for managing \new{corroborative V\&V} in practical applications, as well as acknowledging that \new{corroborative V\&V} is not perfect.
Rather, it is an approach to using V\&V techniques in conjunction with one another to provide a higher degree of confidence that a system will satisfy its requirements.

\new{As we have improved the accuracy of our assets on the basis of the results presented in this paper, we could use the same V\&V techniques to further explore the HRI. For example, it is possible to explore human behaviours that deviate from the typical use case, and incorporate aspects of user uncertainty and variability beyond those used in this paper.}
A reformulation of the system and requirement models under the new conditions might be necessary, as system traits (e.g.,\ failure rates) characterised under one set of constraints will not necessarily hold for other sets of constraints. 
The V\&V engineer should use their judgement in such cases as to whether any prior asset modifications can be generalised to broader scenarios.

The \new{V\&V} efforts towards \new{corroboration} can be helped by limiting (or biasing) the explored region of the HRI state space to seek cases in which the V\&V techniques provide contradictory results.
In our case study, we have employed a probabilistic formulation of the requirements that is relevant to HRI system as non-determinism may arise not only from the environment but also from the robot and the coupling between them (ROS-based robots exhibit high levels of concurrency and run on non-real-time operating systems).
Hence, we can compute conditional probabilities, such as, ``Given  that the robot's gripper fails, what is the probability that the robot warns the user before the object drops?,'' that lead to conditional evidence.

In more complex scenarios, it may become more difficult to identify appropriate modifications to achieve better agreement between assets.
Modifications to system models may be informed by knowledge gained during V\&V.
For example, useful insights may be contributed from systematic risk analyses, such as Fault Tree Analysis or HAZOP (Hazard Operability).
The latter has recently been proposed for use in human-robot interactions to manage the inherent complexity and uncertainty in such systems \citep{guiochetModelBased2013}.

Increased modelling effort is an evident limitation of \new{corroborative V\&V}.
However, the use of multiple V\&V techniques brings savings to this effort.
As our case study demonstrates, early use of more abstract methods allows gradual commitment of resources to more realistic and expensive techniques.
Discrepancies can highlight oversights and areas of uncertainty, informing the judicious use of more expensive techniques (e.g., to characterise the uncertain grip failure rate before proceeding with user experiments in our case study).

For more comprehensive V\&V efforts, Coverage Driven Verification~\citep{araizaillan2016} may be used with \new{corroborative V\&V}, pursuing coverage of the system in a systematic way in simulations or experiments.
Hybrid systems methods~\citep{Julius2007,Kim2006} might also be usefully incorporated into \new{corroborative V\&V}, although reducing entire HRI scenarios to manageable hybrid models is likely to be challenging.

The object handover is only an example of a huge variety of case studies available in the HRI domain.
Nonetheless, it is of uttermost interest in HRI, as close-proximity manipulation tasks may be considered in a plethora of applications, such as manufacture of white goods, cooperative handling and attachment of large sub-components of aeroplane structures in aerospace assemblies, or care of the elderly with early stage dementia by feeding them soup. 

\new{While our approach can be extended to any HRI application, in principle, an awareness of the limitations of each V\&V technique is essential. For example, human behaviour is notoriously difficult to analyse and assess, with open-ended and physically-unconstrained interactions between humans and robots being some of the most difficult problems in HRI research. In complex and nuanced scenarios, we may wish to emphasise the use of experimentation and real-world operations over simulation and formal verification as providing core evidence for corroborative V\&V. However, it is likely that many complex interactions can be broken down into simpler sub-interactions, such as object handover. In these cases, the high levels of efficiency, coverability and precision offered by formal verification and simulation-based testing can be more readily utilised.}

\begin{newsection}
\subsection{Use of Other V\&V Techniques}
\label{s:other}

The corroborative V\&V approach can also make use of other V\&V techniques, as well as their accompanying assets. For example, hardware-in-the-loop experiments allow hardware modules to be used alongside simulated hardware in order to verify the behaviour of those modules~\citep{martin06architecture}. In terms  of abstraction level, hardware-in-the-loop fits between simulation and experimentation, as it makes use of both. Hardware-in-the-loop experiments can verify textual requirements as well as code assertions, and the system model is a combination of the hardware module(s) and simulator. Therefore, we could add hardware-in-the-loop as a V\&V technique within a corroborative V\&V approach (see Figure~\ref{f:hil}).

\begin{figure}[hbt]
\centering
\fcolorbox{white}{white}{\includegraphics[width=1.0\columnwidth]{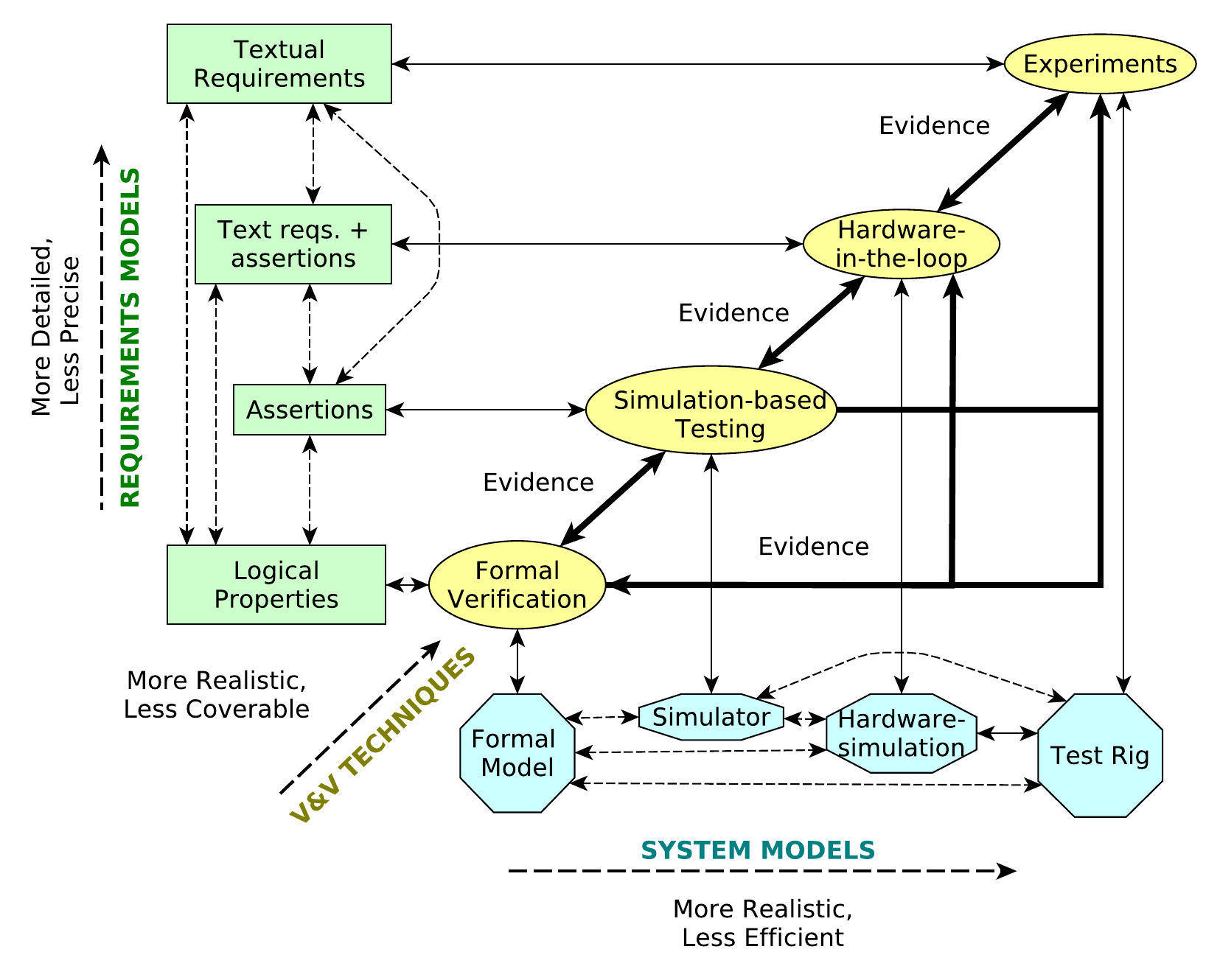}}
\caption{\new{Corroborative V\&V including hardware-in-the-loop.}}
\label{f:hil}
\end{figure}

Of course, we could also expand corroborative V\&V to include operations of the robotic system once it is deployed in the ``real world.'' Once the system is deployed it is being used and operated by its end-users, so naturally, its ``requirement model'' is the end-users' actual requirements, rather than a model captured in natural or formal languages (see Figure~\ref{f:operations}). We could also %
use other V\&V techniques like Coverage-Driven Verification~\citep{araizaillan2015}.

\begin{figure}[hbt]
\centering
\fcolorbox{white}{white}{\includegraphics[width=1.0\columnwidth]{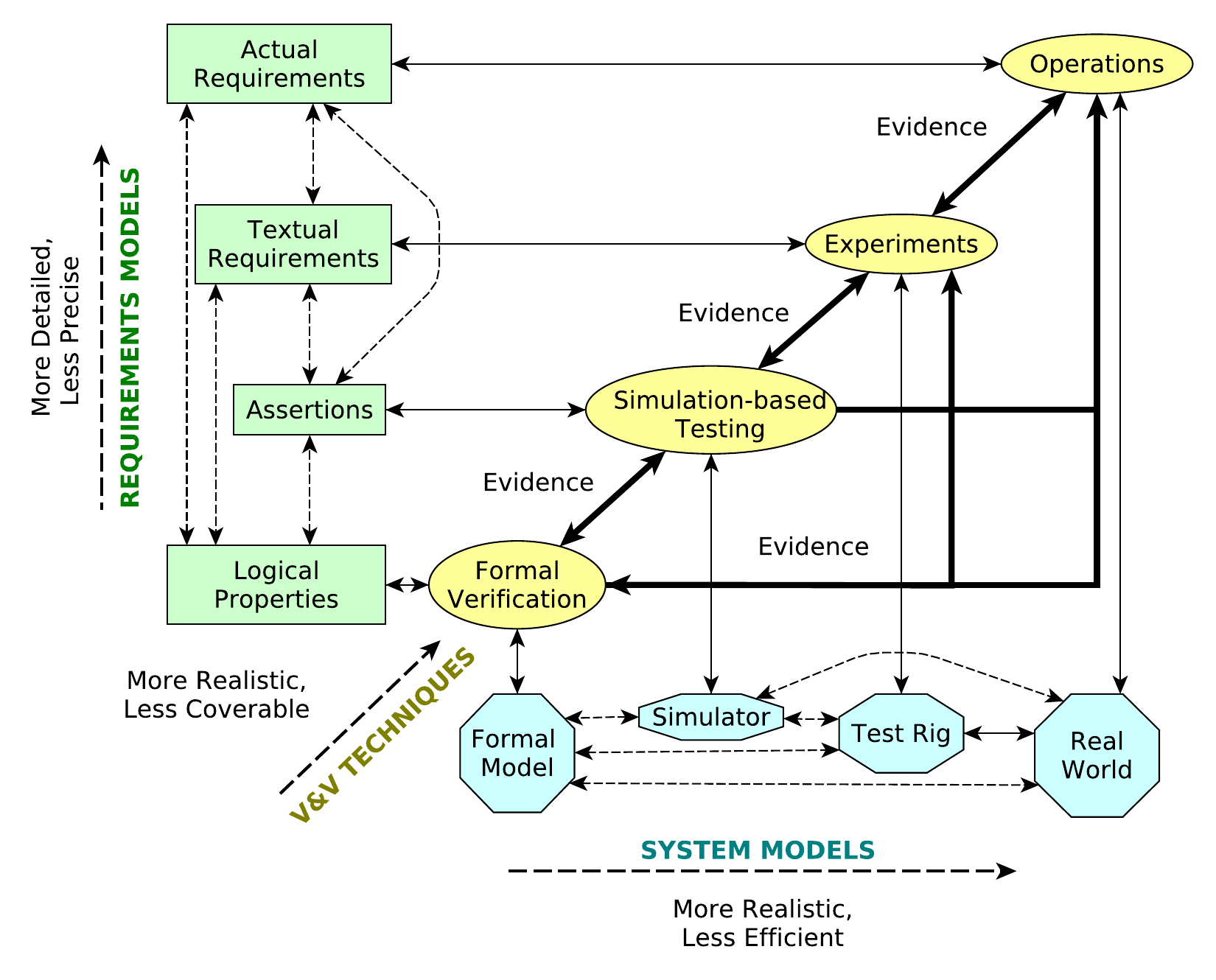}}
\caption{\new{Corroborative V\&V including real-world operations.}}
\label{f:operations}
\end{figure}

The entire corroborative V\&V approach is summarised in Figure~\ref{f:summary}. A set of requirements models (shown in rectangles) are linked to system models (shown in octagons) through a set of V\&V techniques. Information gained from V\&V techniques can be compared with other V\&V techniques, shown by bold arrows. This information can also be fed back to requirements models and system models (i.e., the V\&V assets) in order to refine them to improve accuracy if the V\&V techniques have shown that there is \newer{insufficient} corroboration. The assets can then be refined and compared to one another, before conducting further V\&V until all techniques corroborate one another. Of course, as we have shown in this paper, this is an ideal case, and often full corroboration will not be possible. However, the practice of corroborative V\&V allows and encourages a systematic approach to reaching agreement between V\&V approaches. In turn, this approach to V\&V produces a higher quality of verification and validation than could be achieved by using the individual approaches separately. 

\begin{figure}[hbt]
\centering
\fcolorbox{white}{white}{\includegraphics[width=\columnwidth]{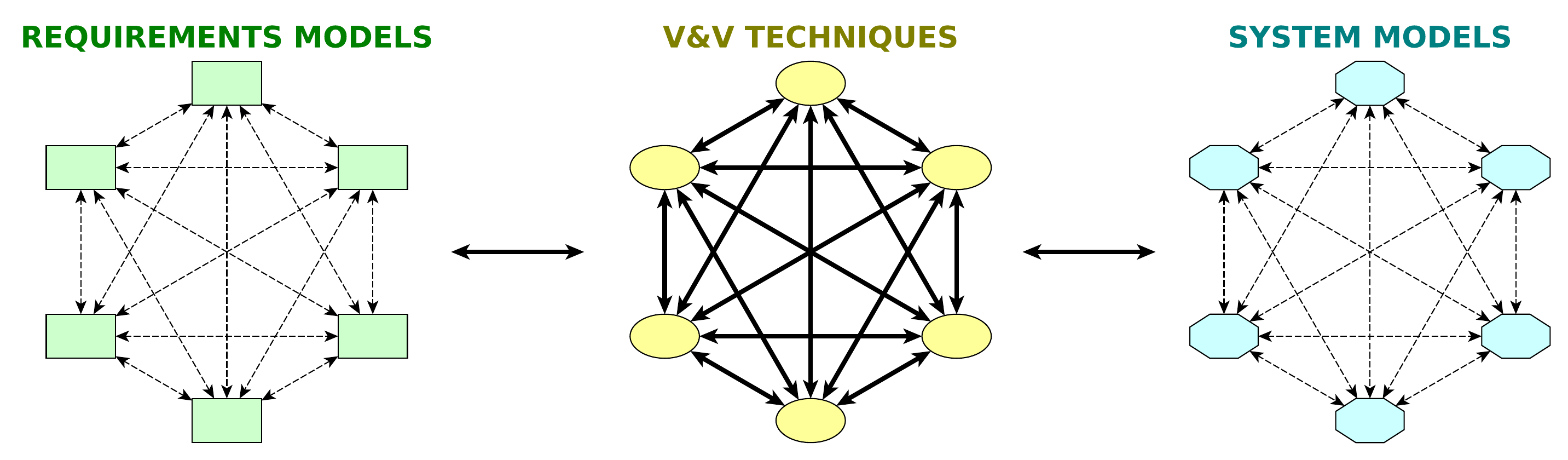}}
\caption{\new{A more abstract view of corroborative V\&V using multiple V\&V techniques.}}
\label{f:summary}
\end{figure}

\end{newsection}
 
\section{\new{Comparison with Other Approaches}}\label{s:relatedwork}

\new{In this paper we have described corroborative V\&V: an approach to V\&V of robotic systems based on combining different V\&V techniques and comparing the evidence generated by them. This was motivated partly by a well-known issue: the use of a single method for V\&V results in a compromise between examining the full state space of a system (in this case, of an HRI) and modelling the system in satisfactory detail.}

\subsection{\new{Formal Methods for V\&V}}

Model checking~\citep{ClarkeMC,Practical11}, a formal method used for V\&V, is exhaustive over the state space of a model but requires abstraction of the full system (e.g., the high-level control algorithms, low-level control and mechanical behaviour, and the code that runs in the robot) into a finite number of states.
For this reason, formal verification can been applied to the analysis of high-level decision-making engines for safety and liveness purposes, exemplified by our previous work in HRI scenarios~\citep{BFS09:HRIshorter,dixon14fridge,webster15toward,gainer17cruton}.
Reasoning and high-level control algorithms have been verified through formal verification and model checking for other kinds of autonomous robots such as ground robots~\citep{doi:10.1177/0278364917733549}, unmanned aircraft~\citep{JACIC13} and multi-robot swarm systems~\citep{Dixon20121429,Konur2012}.
Theorem proving, another formal method, has also been used to verify some of the control code of an autonomous robot~\citep{Walter2010} and multi-robot swarms~\citep{BDF09}, highlighting the same modelling challenges in terms of abstractions versus accuracy and expressivity as in model checking.

\subsection{\new{Simulation for V\&V}}

\new{Although formal models can be run in simulation-mode when model checking is not practical~\citep{Nielsen2014}, dedicated simulators are preferred for robotics V\&V. }
Unlike formal methods, simulation-based testing is not exhaustive and cannot offer proof of requirement satisfaction. However, simulators allow more detailed modelling of the physical and low-level implementation aspects (e.g., sensors or %
joint controllers in the actuators), and the robot's actual control code can be executed for V\&V purposes. This is because simulators do not need to be exhaustive, so computational resources can be used to model systems at a lower-level of abstraction than is seen in formal verification. 
For example, a simulator was built in MATLAB in~\citep{Kirwan2013}, whereas~\citet{Arnold2013} used the Player/Stage 2D simulator and~\citet{Pinho2014} used the SimTwo simulator in combination with the ROS robot software development framework, to test autonomous navigation control algorithms. \newer{Navigation algorithms were also validated in simulation by \citet{robotnavigation17} by using  MORSE, the Modular OpenRobots Simulation Engine\footnote{\url{https://www.openrobots.org/wiki/morse}}.}
In our previous work, we developed simulators in a 3D physical engine, Gazebo, containing models of the robot's joints and continuous motion in space, along with its continuous environment containing objects and humans~\citep{araizaillan2015,araizaillan2016}. 

In other domains, such as microelectronics, both formal methods and simulation-based testing are used, e.g., Electronic Design Automation tools.
Simulation and formal methods have been used in combination to overcome the limitations of model-checking or to provide human-readable evidence of failures that can be observed at runtime. 
Emulating these principles, academic formal analysis tools also offer both model checking and simulation-based testing, such as UPPAAL~\citep{Nielsen2014}, Event-B\footnote{\url{http://www.event-b.org/}}, and the FDR2 tool\footnote{\url{http://www.fsel.com/software.html}}. 
Nonetheless, performing both model checking and testing over the same formal model is disadvantageous when gaining confidence in the resulting V\&V evidence, as these two V\&V techniques are subject to the same modelling and coding errors. 
This problem is highlighted by~\citet{Kirwan2013}, as they crafted a simulator (in MATLAB) and a formal model (in Promela for the SPIN model checker) of their robot's software (an autonomous navigation closed-loop system) independently to overcome the limitations of simulations and model checking and gain confidence in their results. 
~\cite{Intana2013} combined the advantages of simulation and formal verification for wireless sensor networks. Simulations in an environment called MiXiM allowed discovering functional issues in a high fidelity model, whereas formal verification in Event-B was used to provide proofs of requirement satisfaction or violation to strenghten the discoveries during simulation. They do not consider a course of action if the results from simulation contradict the ones from formal verification, as is done in the \new{corroborative V\&V} approach presented in this paper. 

Experiments in real-world scenarios are costly when compared to simulation and formal methods and cannot thoroughly explore the full state space of an HRI scenario. 
Simulation and experimentation can be combined through hybrids of human-in-the-loop and simulation or robot-in-the-loop and simulation as proposed by~\citet{Petters2008}.
However, \new{corroborative V\&V} allows the use of multiple techniques, e.g., formal methods, simulation-based testing, and experiments,  to verify and validate various requirements, eliminating the need to choose between examining the full state space of an HRI and modelling the HRI in satisfactory detail. 

\subsection{\new{Model Validation and Meta-V\&V}}

\new{Our corroborative V\&V approach}  draws on multiple forms of evidence from various V\&V techniques to support a claim. In that sense, corroborative V\&V can be seen as a ``meta-level'' approach to verification and validation, in which V\&V is achieved through a comparison between the  results of different V\&V approaches.

The clear presentation of such arguments, e.g., by Goal Structuring Notation~\citep{kellyGoal2004}, is an important consideration in safety critical systems.
\citet{hawkinsNew2011} describe the importance of separating a safety argument from its accompanying confidence argument, which justifies the sufficiency of confidence in the safety argument.
Similar to our approach, but more limited in terms of variety of V\&V techniques, the claims computed with a new variant of formal analysis, based on models of flows instead of models of states, are validated by experiments in the laboratory~\citep{Lyons2013}. They applied the verification technique to autonomous navigation algorithms for multi-robot missions for Pioneer-3AT robots, but the validation stage only involved one robot. 
An approach to test robotic software through co-simulation was presented in~\citep{Broenink2010}, where formal verification was used to find deadlocks through the FDR2 tool, whereas models of the robot's software and hardware at different levels of abstraction allowed a thorough testing of the discrete and continuous interacting components. These multiple simulators run in a synchronised manner in a co-simulation. They do not consider a course of action when finding discrepancies between the formal analysis and the simulations.

\new{In corroborative V\&V,} we seek agreement between multiple V\&V techniques with respect to particular set of requirements.
Discrepancies may arise due to inaccuracies or errors in one or more of the system models, requirements models, or tools used.
In several previous works, methods have been proposed for improving the models.
For example, formal models are refined iteratively if they produce a spurious property violation after model checking in Counter Example-Guided Abstraction Refinement (CEGAR)~\citep{clarkeCounterexampleGuided2000}. 
An initial detailed model is abstracted to form a simpler upper-approximation for which model checking is tractable.
After encountering a violation of a requirement, that model is iteratively and automatically refined to determine whether the violation is spurious (i.e.,\ does not occur in the more detailed model).
The level of detail that may be accounted for by such techniques remains limited to that which can be formally modelled.
For a system's software, this may extend to the concrete code design, but for complex cyber-physical systems such as HRI, there will typically be important details that cannot be adequately represented.
\new{Corroborative V\&V} may be seen as an approach in the spirit of CEGAR, with greater dependence on human judgement to extend beyond formal modelling and accommodate system models between which absolute agreement may not be achievable.

Many approaches have been proposed to verify and validate requirement models with respect to consistency, completeness, and precision. For example,~\citet{Heitmeyer2007} developed a tool that performs formal verification (both model checking and theorem proving) as well as simulation and even code generation by integrating multiple external tools. Nonetheless, further advantages can be gained when multiple and independently-applied V\&V techniques are combined to gain confidence in their results, as we propose here. 

Frameworks to verify and validate models for simulation tasks with respect to accuracy and validity have been proposed in~\citep{Robinson1997,Sargent2013}. These models are developed from gathering real-world data, and their V\&V continues throughout its simulation use. Dimensions that can be verified are: concept (aspects to be included in the model, such as variables of importance), data (e.g., accuracy, format), timing, control and information flows, and even the code against bugs. Techniques that can be applied for V\&V include animation, comparison against other models, and testing (e.g.,\ stress, sensitivity and historical data comparison). Nonetheless, the authors do not prescribe a methodology with associated tools to achieve model V\&V.

For ROS-based systems, the accuracy of a formal model with respect to the robot's control code may be more rigorously examined by standardising the formal description of common ROS components, e.g., using the ``ROS graph'' formalisation developed in~\cite{aitkenAdaptation2014} to enable automated reconfiguration of ROS systems.  
Recently, this formalisation has been adopted by~\cite{hazimTesting2016} to apply model-checking to the verification of timing properties of ROS processes.
Their approach shows promise for ensuring that the formal model is representative of the system's performance.
Further work is needed to demonstrate whether it can be usefully extended to capture inaccuracies in modelling the environment or the requirements, which may be more challenging to model when considering physical aspects of the system besides timing.

\subsection{\new{Automated Software Tools}}

For both requirement and system models, an approach to ensuring consistency is to generate one model from another by a trusted method.  
Automated tools can help in this process, such as translations from MATLAB/Simulink control models or control code into formal models for model checking~\citep{Xie2004,Meenakshi2006}.
Formal system models can be automatically extracted from real code, as in~\citep{CorbettBSIMCJP,Gallardo2012,Mukhopadhyay2015}, although not many tools are compatible with Python, a popular language for prototyping robotics code.
Logical properties may be automatically converted into automata~\citep{GastinFast2001}, which may then be encoded as a monitor in the form of a finite-state machine.
\citet{huangRosrv2014} introduce \textit{rosmop}, a tool to automatically convert logical properties into monitors for runtime verification of ROS code.
Similar approaches could potentially be adopted in combination with \new{corroborative V\&V} for HRI to increase the level of confidence in the results, although errors could still propagate when transforming one model into another, e.g., inaccuracies in a logical property will propagate to a monitor.

A simulation-based testing process can be improved by using tests that not only stimulate the system but can also find faults, using so-called mutation-based test generation~\citep{Huang2014}. 
A system's safety and liveness requirements model, crucial in a V\&V task, can also be verified for consistency, correctness and completeness, e.g.,\ using a combination of formal methods, static analysis and simulation as in~\citet{Heitmeyer2007}. 
If a system is to be designed and implemented from a requirements model, certified code generators~\citep{Naks2009} and code synthesis (e.g., refinement)~\citep{Ringert2014} can be employed.
However, the validity of the resulting code is dependent on the accurate representation of non-software aspects of the system in the original model, which is especially challenging in the HRI domain.
Furthermore, in practice, robots are commonly designed and built by different interacting teams due to the complexity of the applications.

\smallskip

In summary, various techniques exist to promote correctness in modelling and to bridge different levels of abstraction for V\&V in robotics and other domains.
However, none of these spans the full range of realism and coverability needed to thoroughly verify and validate an HRI system while systematically addressing the possibility for errors to be introduced at any level of abstraction. 
Confidence in the results of these techniques, if used in isolation, is thus limited.
Our proposed approach does not prescribe specific V\&V tools and techniques to be used.
Automatic translation and connections between the used V\&V techniques can be added to the approach, with discretion, to improve confidence or efficiency in the V\&V exercise.
For instance, in our demonstration we exploit model-based test generation and the ROS-Gazebo compatibility as additional links between simulation-based testing and formal methods.
 
\section{Conclusions}\label{s:conclusions}

We presented \new{corroborative V\&V}, a novel approach to the verification and validation of robotic assistants, to help in demonstrating their trustworthiness in the context of human-robot interactions. 
\new{There are a multitude of V\&V techniques, from formal methods like model checking, to various kinds of simulation, hardware-in-the-loop, experimentation and real-world deployment. Naturally, there are trade-offs between different V\&V techniques, e.g., due to abstraction level, ease of modelling and coverability. Furthermore, it is likely that different V\&V techniques may not initially agree on whether a particular system meets a particular requirement. 
\new{Corroborative V\&V} allows us to use the different V\&V techniques together, playing to their individual strengths. Where discrepancies between V\&V techniques are found, \new{corroborative V\&V} can be used to ``iron out'' \newer{these} differences, working towards a situation where the majority of the V\&V techniques are in agreement with respect to a particular set of requirements for a given system.}

Therefore, \new{corroborative V\&V} provides integral assurances on \new{a} robot's safety and functional correctness through the combination of multiple V\&V techniques. %
The use of these techniques provides \new{corroboration} at different degrees of \textit{coverability} (i.e.,\ the exploration of the HRI task) and HRI modelling expressivity, thus overcoming the shortfalls of each technique when applied in isolation.
For example, model checking provides an exhaustive exploration of a system model, but at the cost of system detail, which is often lost in an abstract model. However, in simulation-based testing, we gain high-fidelity detail by running the real software, but we cannot test the whole state space of variables and behaviours.
Also, an iterative process between the different V\&V techniques can be used if the resulting evidence presents discrepancies, refining and improving the \new{assets} (i.e., system and requirement models) to represent the HRI task in a more truthful manner.
This allows a greater level of confidence in the resulting evidence about the safety and functional correctness of the robot.

We demonstrated our \new{corroborative V\&V} approach through a handover task, a safety-critical part of a complex cooperative manufacture scenario, for which we proposed safety and liveness requirements.
We constructed formal models (Probabilistic Timed Automata), a simulator (in the Robot Operating System and Gazebo), and a test rig for the HRI (in the Bristol Robotics Laboratory), as well as temporal logic properties and assertion checkers  from the requirements. 
The V\&V focus starts with a pair of requirements, \textbf{Reqs.\ 1a--b}, for which we sought \new{corroboration} between the three techniques by modifying the formal model and the simulator. We then examined the rest of the requirements, finding previously unknown functional failures in the system. 
Our results showcase the benefits of our approach in terms of thorough exploration of the system under V\&V at different levels of detail and completeness and in terms of gaining confidence in the V\&V results through \new{corroboration}. 

\subsection{Future Work}

We will investigate how the translational potential of our proposed approach can be improved by more explicit evaluations of confidence.
For example,~\citet{guiochetModel2015} summarise various qualitative and quantitative approaches to assessing confidence in V\&V evidence.
They present a quantitative model describing the propagation of confidence through particular argument structures.
The results of our demonstration of \new{corroborative V\&V} constitute an ``alternative argument'' structure, in that separate pieces of evidence can \new{corroborate} each other.
Where one technique provides limited assurance --- e.g., testing covers a limited state-space of a higher fidelity model while model checking covers the full state-space of a lower fidelity model --- this may be accounted for by applying weighting factors to individual pieces of evidence provided by each technique.
For probabilistic traits of a system, statistical techniques such as the modified Wald method~\citep{agrestiApproximate1998} can be used to quantify the uncertainty (confidence intervals) arising from the limited number of tests feasible in simulation or experiment.
However, it should be noted that such confidence intervals do not describe the accuracy of the models themselves.
Hence the implementation of quantitative models of confidence propagation will often rely on informal estimates of the confidence in individual pieces of evidence.

Finally, we intend to apply corroborative V\&V to a broader collaborative manufacturing task, of which handover may be a subcomponent.

\section*{Acknowledgments}
\newer{This work was supported by the EPSRC under the the FAIR-SPACE (EP/R026092/1), RAIN (EP/R026084/1)
and ORCA (EP/R026173/1) RAI Hubs, the ``Science of Sensor Systems Software'' programme grant (EP/N007565/1), the ``Trustworthy Robotic Assistants'' project (EP/K006320/1, EP/K006\-193/1 and EP/K0062\-23/1), and the RIVERAS project (EP/J01205X/1). The authors thank Kyriakos Georgiou for his support with data management.}

\end{document}